\documentclass[review]{elsarticle}

\usepackage[a4paper, total={6in, 9in}]{geometry}

\usepackage{lineno,hyperref}
\modulolinenumbers[5]

\usepackage[utf8]{inputenc}

\usepackage{graphicx}
\usepackage[ruled]{algorithm2e}
\usepackage{amssymb}
\usepackage{amsmath}
\usepackage{multirow}
\usepackage{adjustbox}	
\usepackage{booktabs}
\usepackage{url}
\usepackage[acronym,nonumberlist]{glossaries}
\usepackage{arydshln}
\usepackage[dvipsnames]{xcolor}
\usepackage{subcaption}
\usepackage[inline]{enumitem}
\usepackage{float}
\newacronym{mir}{MIR}{Music Information Retrieval}
\newacronym{mtd}{MTD}{Musical Theme Dataset}
\newacronym{omr}{OMR}{Optical Music Recognition}
\newacronym{amt}{AMT}{Automatic Music Transcription}
\newacronym{mlp}{MLP}{Multi-Layer Perceptron}
\newacronym{peft}{PEFT}{Parameter-Efficient Fine-Tuning}
\newacronym{lora}{\textsc{LoRA}}{Low-Rank Adaptation}
\newacronym{qlora}{\textsc{QLoRA}}{Quantized Low-Rank Adaptation}
\newacronym{lp}{LP}{linear probing}
\newacronym{pp}{pp}{percentage points}

\newcommand{\mert}{\textsc{MERT}}
\newcommand{\vit}{\textsc{ViT}}
\newcommand{\roberta}{\textsc{RoBERTa}}
\newcommand{\triscore}{\textsc{TriScore}}
\newcommand{\primus}{\textsc{Primus}}
\newcommand{\KERN}{\texttt{kern}}
\newcommand{\dit}{\textsc{DiT}}
\newcommand{\knn}{$k$NN}
\newcommand{\up}[1]{\textcolor{ForestGreen!60!black}{$\uparrow$#1pp}}
\newcommand{\down}[1]{\textcolor{red!60!black}{$\downarrow$#1pp}}

\graphicspath{{figs/}}
\DeclareGraphicsExtensions{.pdf,.png,.jpg}

\journal{Journal of ...}

\begin{document}
\begin{frontmatter}

\title{Parameter-Efficient Fine-Tuning of Transformer Encoders for Audio–Image–Symbolic Music Retrieval}

\author[add1]{Eliseo Fuentes-Martinez\corref{cor1}}
\cortext[cor1]{Corresponding author}
\ead{eliseo.fuentes@ua.es}

\author[add1]{Antonio Hidalgo-Centeno}
\ead{a.hidalgo@ua.es}

\author[add1]{Jorge Calvo-Zaragoza}
\ead{jcalvo@dlsi.ua.es}

\author[add1]{Antonio Javier Gallego}
\ead{jgallego@dlsi.ua.es}

\address[add1]{Universidad de Alicante, C/ San Vicente del Raspeig s/n, San Vicente del Raspeig (Alicante) - 03690, Spain}

\begin{abstract}
Score-based multimodal music retrieval requires representations that are both discriminative within each modality and compatible across audio recordings, sheet music images, and symbolic scores. This paper investigates the effectiveness of parameter-efficient fine-tuning (PEFT) for adapting pretrained music encoders and examines whether modality-specific improvements translate into better multimodal representations. We compare full fine-tuning, LoRA, and QLoRA across nine pretrained audio, image, and symbolic score encoders on \triscore{} and the Musical Theme Dataset. Encoders are first evaluated and selected through supervised unimodal classification, using frozen-encoder linear probing as a reference. We then introduce a controlled $2\times2$ design crossing encoder adaptation with task-specific shared projection, comparing frozen and adapted representations before and after projection. LoRA or QLoRA is selected in 14 of 18 modality--target combinations, frequently matching or outperforming full fine-tuning while optimizing substantially fewer parameters. However, these unimodal gains do not transfer uniformly to multimodal classification. Adaptation and projection are complementary for composer classification on \triscore{}, largely redundant for \triscore{} instrument and the Musical Theme Dataset instrument and ensemble tasks, and partially interfering for composer classification on the Musical Theme Dataset. The final adapted-and-projected title representation on \triscore{} supports piece-level retrieval, reaching average Top-5 and Top-10 accuracies of 55.32 and 64.68, respectively. Cross-modal exclusion and neighborhood analyses further reveal strong modality-specific local structure in the projected spaces. Overall, PEFT provides an effective and computationally efficient alternative to full fine-tuning, but improved unimodal discrimination alone does not guarantee better multimodal organization.
\end{abstract}


\end{frontmatter}


\section{Introduction}
\label{sec:introduction}

Multimodal approaches to score-based music processing have enabled applications such as piece identification~\cite{Dorfer-2018}, content-based retrieval~\cite{carvalho2023self,pasini2025music2latent2}, cross-modal alignment~\cite{stefan_balke_2019_3527782}, and unified cross-modal transcription~\cite{jung2025unifiedcrossmodaltranslationscore}. In these settings, audio recordings, sheet music images, and notation-level symbolic scores provide complementary views of the same musical content. Audio captures its acoustic realization, symbolic scores encode structured musical notation, and score images preserve the visual form in which music is commonly read, stored, and processed. Learning representations that connect these heterogeneous sources is therefore a central problem in score-based \gls{mir}.

Earlier approaches in the field often relied on intermediate representations, such as chroma features or bootleg score representations, to bridge the gap between heterogeneous music modalities~\cite{fremerey2009_sheet_audio_identification, zalkow2020_weakelyscoreaudioretrieval, tsai2020_cellphonemidi}. However, these pipelines depend on preprocessing stages and hand-crafted representations that may introduce errors and often require substantial domain expertise. This motivated a shift toward deep learning approaches capable of learning task-relevant representations more directly from data~\cite{dorfer_2016, Dorfer-2018}. More recently, the increasing availability of large pretrained models has enabled multimodal systems to build on rich representations extracted from modality-specific encoders, which can subsequently be combined or mapped into a shared representation space~\cite{qin2024, wu2023clampcontrastivelanguagemusicpretraining, wu2025_clamp3}.

Pretrained encoders can either be kept frozen or adapted to the target domain. Frozen feature extraction is computationally convenient and preserves the general knowledge acquired during pretraining, but it may leave relevant musical structure underrepresented. Fine-tuning offers greater adaptation capacity by updating the pretrained representation itself, either by modifying only a subset of the encoder layers or by optimizing the entire backbone. In the latter case, full fine-tuning provides maximum flexibility but requires updating and storing a separate set of model parameters for every encoder, modality, dataset, and task. This cost becomes particularly important in multimodal systems, where several large transformer backbones may need to be adapted and maintained simultaneously.

\gls{peft} methods provide an intermediate alternative by updating only a small subset of model parameters~\cite{han2024parameterefficient}. Among them, \gls{lora}~\cite{hu2022_lora} freezes the pretrained backbone and introduces trainable low-rank updates into selected transformer layers. \gls{qlora}~\cite{dettmers2023qlora} further reduces memory requirements by combining low-rank adaptation with a quantized frozen backbone. Although these techniques have been widely studied in natural language processing and, increasingly, in vision and multimodal models, their effectiveness across the heterogeneous encoders used in score-based multimodal \gls{mir} remains underexplored. This setting is particularly relevant for \gls{peft}, as multimodal systems may require separate adaptations of several large backbones across modalities, datasets, and supervision targets.

This scenario raises two related questions. First, it is unclear whether parameter-efficient adaptation can match or outperform full fine-tuning across audio, image, and symbolic score encoders while substantially reducing the number of trainable parameters. It is also unclear whether these adaptation strategies affect all score-related modalities equally. Music-pretrained audio models may already encode useful domain knowledge and therefore require relatively limited adaptation, whereas image and symbolic encoders transferred from less closely related domains may benefit more strongly from task-specific supervision. The effectiveness of adaptation may therefore depend on the modality, the pretrained encoder, and the supervision target.

Second, and more fundamentally, improved unimodal discrimination does not necessarily imply improved multimodal compatibility. Fine-tuning may increasingly specialize an encoder toward target-discriminative cues that are specific to its input modality, potentially at the cost of losing some of the more general structure inherited from pretraining. Such cues may be highly predictive within one modality but weakly represented, represented differently, or entirely absent in the others. A modality-specific encoder can therefore become substantially better at predicting a target label without producing a representation that is easier to integrate into a common multimodal space. Conversely, a shared projection may recover useful category-level organization from frozen representations even when no task-specific encoder adaptation has been applied.

This distinction motivates two central questions. First, to what extent is task-specific encoder adaptation beneficial over using frozen pretrained representations, and, when adaptation is beneficial, can parameter-efficient methods such as LoRA and QLoRA provide a competitive alternative to full fine-tuning across score-related modalities? Second, to what extent do the resulting unimodal improvements translate into better multimodal representations? The first question requires a systematic comparison of frozen pretrained encoders, full fine-tuning, and parameter-efficient adaptation across modalities, encoders, and supervision targets. The second requires separating two transformations that are often evaluated jointly: adapting each modality encoder to a supervision target and learning a shared projection across modalities. Without this separation, improvements in the final representation cannot be attributed to encoder adaptation, the projection mechanism, or their interaction.

\begin{figure}[t]
    \centering
    \includegraphics[width=.7\linewidth]{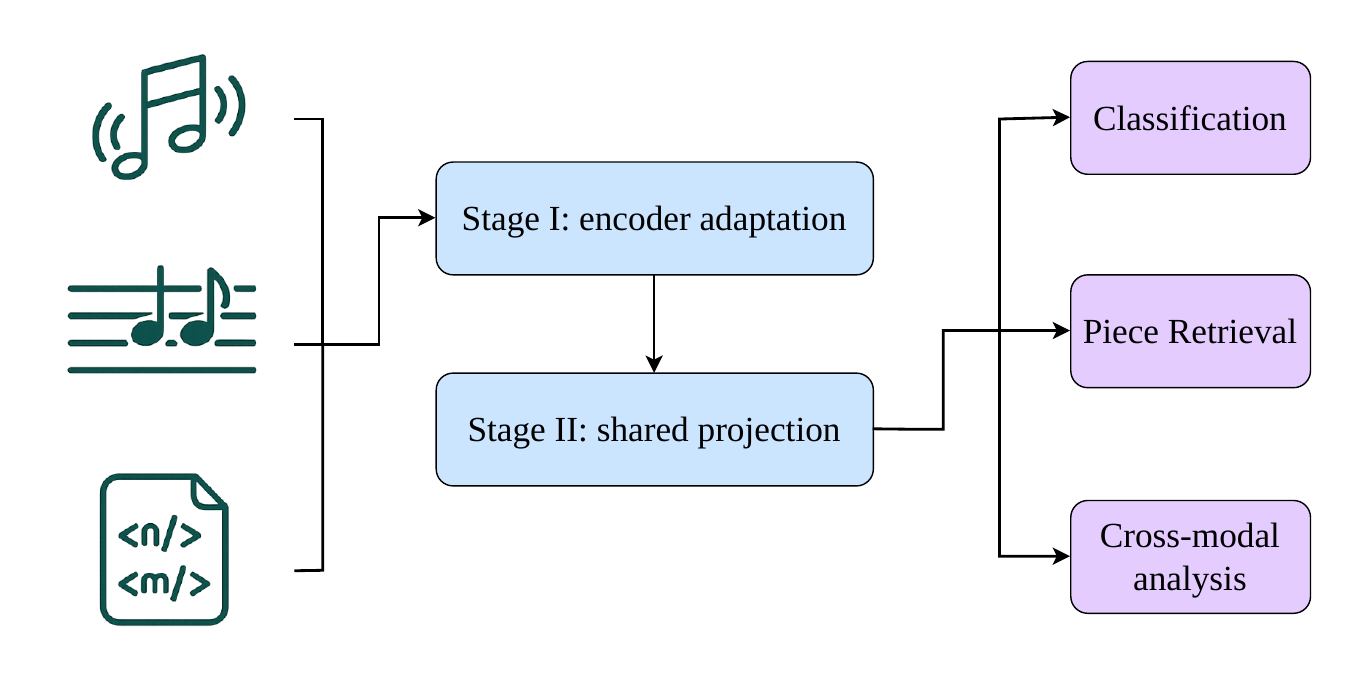}
    \caption{High-level overview of the proposed evaluation pipeline. In Stage~I, modality-specific audio, image, and symbolic score encoders are adapted and selected through unimodal supervision. In Stage~II, their representations are mapped through a task-specific shared projection and evaluated in three downstream settings: category-level classification, piece-level retrieval, and cross-modal analysis.}
    \label{fig:overview}
\end{figure}

We investigate these questions by comparing full fine-tuning, \gls{lora}, and \gls{qlora} across nine pretrained transformer encoders, with three alternatives for each of the audio, image, and symbolic score modalities. Experiments are conducted on two complementary score-based multimodal datasets, \triscore{}~\cite{triscore} and \gls{mtd}~\cite{ZalkowBAM20_MTD_TISMIR}, which differ in annotation structure, label granularity, cross-modal equivalence, and task difficulty. Figure~\ref{fig:overview} provides a high-level overview of the evaluation pipeline.

The evaluation follows a two-stage protocol. In Stage~I, each encoder is assessed independently through supervised unimodal classification, using frozen-encoder \gls{lp} as a reference and full fine-tuning, \gls{lora}, and \gls{qlora} as adaptation strategies. Stage~II examines whether the gains obtained through unimodal adaptation transfer to the multimodal setting. For each dataset and category-level target, the selected adapted encoders are compared with their original frozen counterparts in a controlled $2\times2$ design crossing \emph{encoder adaptation} (frozen vs.\ adapted) with \emph{shared projection} (raw vs.\ projected). This design separates the effects of adaptation, projection, and their interaction. The resulting representations are evaluated in three complementary downstream settings: category-level classification, piece-level retrieval, and cross-modal analysis through exclusion and neighborhood composition. Piece-level retrieval is evaluated separately using the final title-specific adapted-and-projected representation, whereas the cross-modal analyses examine whether category-level coordination is accompanied by local integration across modalities.

The main contributions of this work are as follows:

\begin{itemize}
    \item We provide a systematic evaluation of frozen pretrained representations, full fine-tuning, \gls{lora}, and \gls{qlora} across nine audio, image, and symbolic score encoders and two multimodal music datasets, characterizing the trade-off between predictive performance and the number of trainable parameters.
    
    \item We introduce a controlled $2\times2$ evaluation crossing encoder adaptation with task-specific shared projection, allowing their individual and joint effects on category-level multimodal classification to be distinguished.
    
    \item We evaluate the resulting representations across complementary downstream settings, including category-level classification, piece-level retrieval using the final title-specific representation, and cross-modal exclusion and neighborhood analyses that characterize the local organization of the shared spaces.
    
    \item We show that improvements in unimodal discrimination do not necessarily translate into better multimodal representations: encoder adaptation and shared projection can have complementary, redundant, or interfering effects depending on the dataset and supervision target.
\end{itemize}

The remainder of this paper is organized as follows. Section~\ref{sec:related_works} reviews related work on score-based multimodal \gls{mir}, pretrained encoders, and parameter-efficient fine-tuning. Section~\ref{sec:materials_methods} presents the proposed two-stage methodology, and Section~\ref{sec:experimental_setup} describes the datasets and experimental setup. Section~\ref{sec:results} presents and discusses the unimodal adaptation, multimodal classification, retrieval, and cross-modal analysis results. Finally, Section~\ref{sec:conclusions} summarizes the main conclusions, limitations, and future directions.

\section{Related Works}
\label{sec:related_works}

Multimodality is intrinsic to music, since the same work may be represented through audio recordings, symbolic notation, sheet music images, metadata, lyrics, video, or combinations of these sources. Each representation may capture different and potentially complementary aspects of the musical content. This has motivated a broad range of multimodal approaches in \gls{mir}, including audio--language learning~\cite{guinot2025_slap,ilaria2022_audiolanguagelearning}, lyrics-aware representation learning~\cite{musictm}, metadata-enriched classification~\cite{tabaza2024representationlearning}, image-based retrieval~\cite{luis_carvalho_2023_10265385}, score following~\cite{henkel2021}, and video-based performance analysis~\cite{keunwoochoi2021_listen_read_identify,kim2025_pianovam}. Comprehensive reviews of multimodal \gls{mir} and multimodal music datasets are provided by Christodoulou et al.~\cite{christodoulou2024_multimodaldatasets}, M\"uller et al.~\cite{muller_fundamentals}, and Simonetta et al.~\cite{simonetta2019}.

Within this broader field, the present study concerns \emph{score-based} multimodal representation learning, where audio recordings, notation-level symbolic scores, and sheet music images are treated as related manifestations of musical content. The following subsections review research on cross-modal learning for these representations, the pretrained encoders available for each modality, full and parameter-efficient adaptation methods, and the multimodal datasets used to evaluate them.

\subsection{Score-based multimodal MIR}
\label{sec:related_works:multimodal_mir}

Score-based multimodal \gls{mir} investigates relationships among different representations of notated music, most commonly performance audio, sheet music images, and symbolic score encodings. Earlier systems frequently relied on intermediate representations obtained through \gls{omr}~\cite{Castellanos2025omr} or \gls{amt}~\cite{Benetos2019amt}, supporting applications such as audio--score alignment, score following, and navigation~\cite{thomas2012_LinkingSheetMusic}. More recent representation-learning approaches instead model correspondences directly between heterogeneous inputs~\cite{Dorfer-2018}.

A substantial part of this literature has focused on audio--sheet music retrieval. Dorfer et al.~\cite{dorfer_2016,Dorfer-2018} proposed neural embedding models for matching short audio excerpts with sheet music snippets, establishing an influential basis for learned audio--score retrieval. Later studies investigated more demanding retrieval conditions, including tempo variation~\cite{stefan_balke_2019_3527782}, temporal dependencies and alignment constraints~\cite{carvalho2021exploiting,luis_carvalho_2023_10265385, bukey2026_LocalMultimodalMusic}, and retrieval from less controlled performance data~\cite{carvalho2023self}. Collectively, these works demonstrate that learned representations can connect acoustic performances and visual notation without requiring a complete symbolic transcription at retrieval time. However, most operate on a pair of modalities rather than on a representation space shared by audio, image, and symbolic score data.

Symbolic score representations have received comparatively less attention in multimodal retrieval. In much of the \gls{mir} literature, the term \emph{symbolic music} refers to MIDI-like piano-roll or event-based formats. However, notation-level encodings such as MusicXML, MEI, or \KERN{} represent additional score structure and notational information, and are more directly related to the corresponding visual score. Systems such as CLaMP~\cite{wu2023clampcontrastivelanguagemusicpretraining, wu2025_clamp3} incorporate symbolic music into contrastive retrieval frameworks, while other studies investigate multimodal pretraining or translation involving score-related representations~\cite{jung2025unifiedcrossmodaltranslationscore,qin2024}. Nevertheless, these approaches generally emphasize text--music relationships, pairwise correspondence, or cross-modal generation, rather than retrieval within a unified space containing audio, notation-level symbolic scores, and sheet music images.

Unified score-based representation spaces have only recently been studied. Hidalgo-Centeno et al.~\cite{triscore} evaluated frozen pretrained encoders for audio, image, and symbolic score inputs and mapped their outputs through a shared lightweight projection. The resulting space supported classification and piece retrieval, while its local neighborhoods remained strongly organized by modality. This finding illustrates an important distinction for multimodal retrieval: a representation may exhibit category-level consistency across modalities without producing fully mixed instance-level neighborhoods. However, it remains unclear whether improvements obtained by adapting the modality-specific encoders lead to better organization in the resulting shared multimodal space.


\subsection{Pretrained encoders for audio, image, and symbolic music representations}
\label{sec:related_works:pretrained_encoders}

The behavior of a multimodal system depends partly on the representations supplied by its modality-specific encoders. However, the availability and domain specificity of pretrained models differ considerably across audio, visual, and symbolic score data.

In audio-based \gls{mir}, large-scale self-supervised models have become increasingly important. \mert{}~\cite{li2024mertacousticmusicunderstanding} is explicitly pretrained for music understanding and is intended to capture acoustic and musical properties from large audio collections. Other transformer encoders originate from broader audio domains. Wav2Vec2~\cite{quarta2024_wav2vec2_audioset} was initially developed within speech representation learning, although variants trained on AudioSet include music and environmental sounds. Whisper~\cite{radford2022_whisper} is primarily a multilingual speech model, but its scale and acoustic pretraining make its encoder potentially useful beyond speech recognition~\cite{ghosh2025_audioflamingo3}. These models therefore provide different combinations of domain specificity, scale, and pretraining objectives.

Music-specific pretraining is less readily available for sheet music images. General-purpose vision transformers such as \vit{}~\cite{vit} provide broadly transferable visual representations and have been used in multimodal work involving score images~\cite{qin2024}. Document-oriented architectures such as \dit{}~\cite{peebles2023_dit} may be better suited to structured page layouts, while vision--language models such as PaliGemma~\cite{steiner2024_paligemma2familyversatile} provide visual representations learned as part of a larger multimodal architecture. Despite these differences, none of these models is primarily pretrained to represent the specific graphical syntax of conventional music notation. Their suitability for sheet music retrieval may therefore depend substantially on downstream adaptation.

Notation-level symbolic scores pose a related but structurally distinct challenge. Although several transformer architectures have been developed for symbolic music generation and understanding, many target MIDI-like event sequences rather than score encodings such as \KERN{}. General-purpose language encoders such as \roberta{}~\cite{liu2019robertarobustlyoptimizedbert} and SentenceTransformer~\cite{reimers2019_sentence} can process tokenized score sequences, while GraphCodeBERT~\cite{guo2021_graphcodebertpretrainingcoderepresentations} provides a code-oriented alternative that may be suitable for the structured and partly syntactic properties of symbolic notation. However, their pretraining corpora and objectives remain substantially different from those of notation-level music data.

These differences create an asymmetric representation setting. Audio encoders may already contain music-specific acoustic knowledge, whereas visual and symbolic encoders often transfer from document, image, language, or source-code domains. Accordingly, neither the benefit of adaptation nor the relative effectiveness of full and parameter-efficient fine-tuning can be assumed to be uniform across modalities. Both may depend on the relationship between the pretraining domain, the structural properties of the input representation, and the downstream supervision target.

\subsection{Full and parameter-efficient fine-tuning}
\label{sec:related_works:finetuning}

Fine-tuning is a standard mechanism for adapting pretrained transformer representations to a target domain or task. Under full fine-tuning, all backbone parameters are updated together with the task-specific output head. This provides substantial adaptation capacity, but requires optimizing and storing a complete model variant for each experimental condition. The cost is especially relevant in multimodal systems, which may contain a separate large encoder for every modality.

\gls{peft} methods reduce this requirement by updating only a restricted set of parameters while leaving most of the pretrained backbone frozen~\cite{han2024parameterefficient}. Adapter-based strategies introduce small trainable components into transformer layers, allowing a common backbone to support several task- or domain-specific variants. Their lower parameter requirements are attractive when multiple encoders, modalities, and supervision targets must be considered.

Among \gls{peft} approaches, \gls{lora}~\cite{hu2022_lora} represents weight updates through trainable low-rank matrices inserted into selected layers of a frozen model. \gls{qlora}~\cite{dettmers2023qlora} combines low-rank adaptation with a quantized frozen backbone, further reducing the memory required to store the base model. Their parameter and memory efficiency make them particularly attractive for multimodal systems, where several modality-specific backbones may need to be adapted separately for multiple datasets and supervision targets.

However, the relative effectiveness of full fine-tuning, \gls{lora}, and \gls{qlora} has not been systematically established across the heterogeneous encoders used in score-based \gls{mir}. Their performance may depend on model architecture, pretraining domain, modality, supervision target, and adaptation hyperparameters. Moreover, matching or exceeding full fine-tuning in a unimodal task does not ensure that the adapted representation will be more compatible with the other modalities. A complete evaluation must therefore consider both parameter-efficient unimodal adaptation and the extent to which its gains remain beneficial before and after learning a shared multimodal projection.

\subsection{Score-based multimodal datasets}
\label{sec:related_works:datasets}

Research on score-based multimodal retrieval is constrained by the limited availability of datasets containing aligned audio, symbolic scores, and sheet music images. Constructing such resources requires collecting or generating several representations of the same musical material, maintaining reliable correspondence between them, and providing annotations suitable for classification or retrieval.

Several existing datasets partially cover these requirements. MSMD~\cite{Dorfer-2018} contains aligned sheet music images and audio/MIDI representations of classical piano music and has been widely used for audio--sheet retrieval. MusicTM~\cite{musictm} combines sheet music images, lyrics, and synthesized audio, supporting multimodal learning involving notation and language. \primus{}~\cite{ALFAROCONTRERAS2023119491}, originally developed for \gls{omr}, has also been used in combined \gls{amt} and \gls{omr} settings, providing aligned audio, image, and symbolic representations. However, its original annotations were not designed for retrieval-oriented metadata tasks.

The \gls{mtd}~\cite{ZalkowBAM20_MTD_TISMIR} contains more than 2,000 Western classical themes with symbolic encodings, sheet music images, audio excerpts, and manual alignments. It provides a valuable setting for theme-level cross-modal analysis, although the modalities are not fully equivalent in informational content. The symbolic and visual representations primarily describe short monophonic themes, whereas the associated recordings may also contain harmony, instrumentation, texture, and surrounding musical context.

\triscore{}~\cite{triscore}, derived from MUSCUTS and MUSCAT~\cite{muscat2024}, provides excerpt-level triplets consisting of audio recordings, \KERN{} symbolic scores, and rendered score images. Its annotations include composer, instrument, and piece title, enabling both category-level classification and piece-level retrieval. Compared with \gls{mtd}, its modalities represent more closely corresponding musical excerpts, providing a more controlled alignment setting.

We use \triscore{} and \gls{mtd} as complementary evaluation datasets. Their differences in cross-modal equivalence, annotation structure, and label distributions make it possible to examine whether the transfer of unimodal adaptation gains depends on the informational compatibility of the modalities being combined.

\subsection{Positioning of this work}
\label{sec:related_works:positioning}

The literature leaves two related questions unresolved. First, it remains unclear whether parameter-efficient fine-tuning can provide a competitive alternative to full fine-tuning across the heterogeneous audio, image, and notation-level symbolic encoders used in score-based \gls{mir}. Existing models differ substantially in architecture, pretraining domain, and compatibility with musical inputs, but full fine-tuning, \gls{lora}, and \gls{qlora} have not been compared across these modalities under a common experimental protocol. Second, stronger unimodal discrimination does not necessarily imply better multimodal organization, and it remains unclear whether adaptation gains persist when modality-specific representations are integrated into a shared space.

We address the first question through dataset-, modality-, and target-specific comparisons of frozen-encoder \gls{lp}, full fine-tuning, \gls{lora}, and \gls{qlora}, considering both classification performance and the number of trainable parameters. We address the second through a controlled $2\times2$ evaluation crossing encoder adaptation with task-specific shared projection. This design distinguishes the individual and joint effects of the two transformations and reveals whether they behave complementarily, redundantly, or in an interfering manner.

\section{Methodology}
\label{sec:materials_methods}

\begin{figure*}[t]
    \centering
    \includegraphics[width=\textwidth]{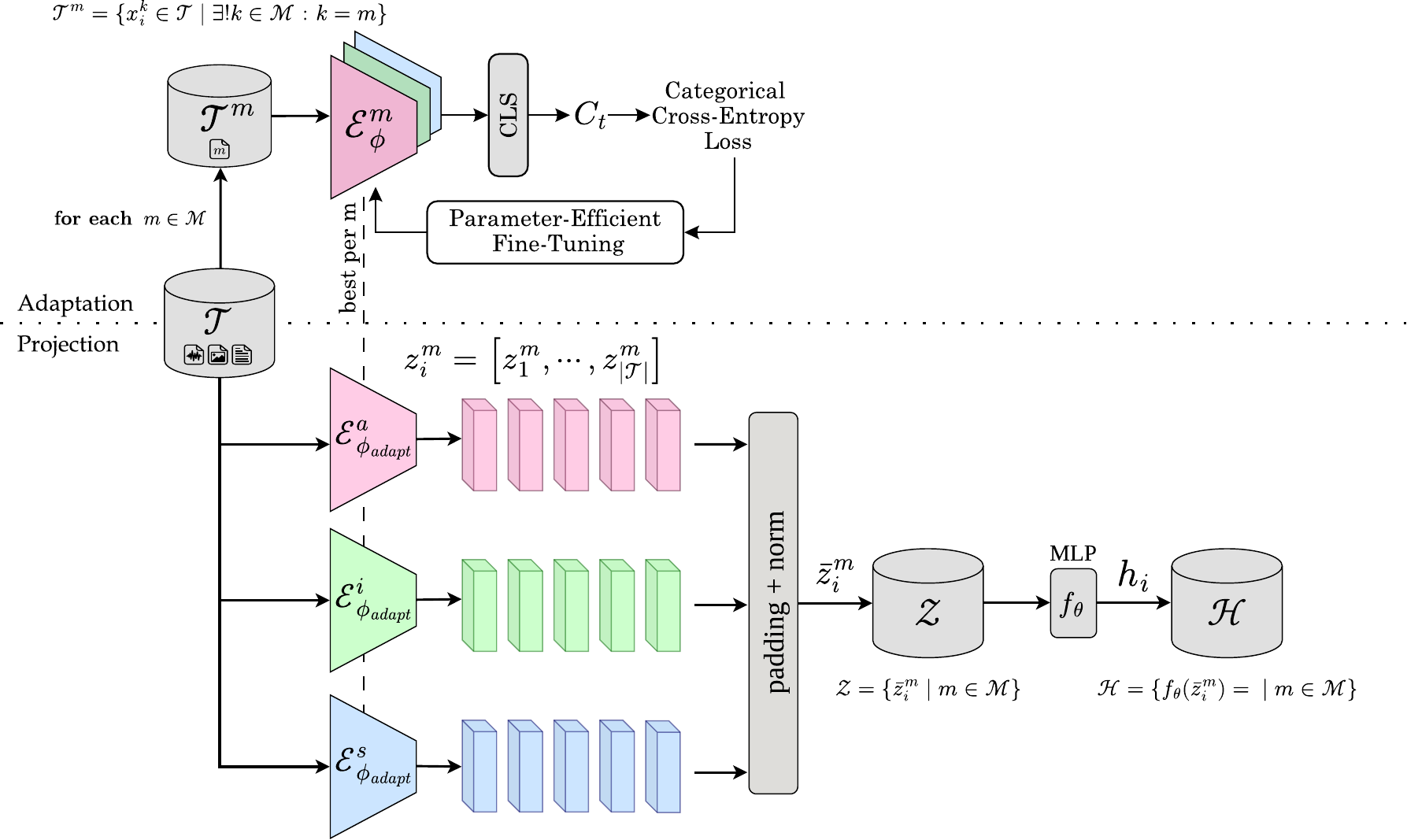}
    \caption{Overview of the proposed two-stage methodology. Stage~I adapts each candidate pretrained encoder $\mathcal{E}$ independently through supervised unimodal classification. Frozen-encoder linear probing is used as a reference, while full fine-tuning, LoRA, and QLoRA are compared as adaptation strategies; the best-performing encoder--adaptation configuration is selected using validation macro F1. Stage~II extracts representations from both the selected adapted encoder $\mathcal{E}_{\phi_{adapt}}^{m}$ for each modality $m$.}
    \label{fig:methodoloy}
\end{figure*}

Figure~\ref{fig:methodoloy} summarizes the proposed methodology. Let $\mathcal{M}=\{\mathrm{audio},\mathrm{image},\mathrm{symbolic}\}$ denote the set of modalities, and let $x_i^m$ be the input associated with item $i$ in modality $m\in\mathcal{M}$. Each experiment is defined for a dataset $\mathcal{D}$ and a supervision target $t$, with label $y_i^t$. The method is divided into two stages. Stage~I adapts candidate pretrained encoders independently for each modality, dataset, and target. Stage~II then performs a multimodal evaluation where the extracted representations from the adapted encoders are mapped to a joint representation.

The two stages address different questions. Stage~I compares the discriminative capacity and computational cost of frozen pretrained representations, full fine-tuning, \gls{lora}, and \gls{qlora}, while Stage~II evaluates the effectiveness of the shared projection from the extracted unimodal representations. The concrete pretrained encoders considered for each modality, datasets, target labels, optimization settings, and evaluation protocols are specified in Section~\ref{sec:experimental_setup}.

\subsection{Stage I: Unimodal Encoder Adaptation}
\label{sec:materials_methods:ft}

The first stage evaluates the target-specific discriminability of each candidate modality encoder and compares alternative mechanisms for adapting its pretrained parameters. Each encoder is trained independently over a subset of the training set $\mathcal{T}$ where only the samples corresponding to that specific modality are selected, i.e.,

\begin{equation}
    \mathcal{T}^m = \{ x_i^k \in \mathcal{T} \mid \exists! k \in \mathcal{M} \text{ : } k = m \}
\end{equation}

The adaptation strategy considers a unimodal pretrained encoder $\mathcal{E}_{{\phi_0}}^{m}$ for modality $m$, where $\phi_0$ denotes the encoder parameters obtained through large-scale pretraining. Let $z_i^{m}=\mathcal{E}_{\phi}^{m}(x_i^m)$ be its pooled representation for input $x_i^m$. A target-specific linear classifier maps this representation to a probability distribution over the $C_t$ classes of target $t$; the classifier and the encoder adaptation parameters are optimized using cross-entropy loss.  Four encoder adaptation strategies are considered, where each adaptation specifies how the encoder parameters $\phi_0$ participate in the training over the training set $\mathcal{T}$, yielding the adapted encoder parameters $\phi_{adapt}$:

\begin{itemize}
    \item In \emph{linear probing}, the pretrained backbone remains frozen and only the linear classifier is optimized. This condition provides a reference estimate of the target-relevant information already accessible in the original representation. Thus, the weights of the pretrained encoder remain unchanged, i.e., $\phi=\phi_0$.
    
    \item In \emph{full fine-tuning}, all encoder and classifier parameters are updated, i.e. $\phi=\phi^*$. This provides the highest adaptation capacity but also requires a complete task-specific copy of the backbone and the largest optimization memory.
    
    \item In \gls{lora}, the pretrained backbone remains frozen and trainable low-rank updates are introduced into selected linear transformations. For a pretrained weight matrix $\mathbf{W}_0\in\mathbb{R}^{d_{\mathrm{out}}\times d_{\mathrm{in}}}$, \gls{lora} represents the adapted transformation as
        \begin{equation}
            \mathbf{W} = \mathbf{W}_0+\Delta\mathbf{W},
            \qquad
            \Delta\mathbf{W} = \frac{\alpha}{r}\mathbf{B}\mathbf{A},
            \qquad
            \mathbf{W} \in \phi
            \label{eq:lora_update} 
        \end{equation}
    where $\mathbf{A}\in\mathbb{R}^{r\times d_{\mathrm{in}}}$ and $\mathbf{B}\in\mathbb{R}^{d_{\mathrm{out}}\times r}$ are trainable matrices, $r\ll\min(d_{\mathrm{in}},d_{\mathrm{out}})$ is the adapter rank, and $\alpha$ controls the magnitude of the update~\cite{hu2022_lora}. 
    
    \item In \gls{qlora}, the same low-rank adaptation mechanism is applied while the frozen backbone is stored in quantized form, further reducing its memory footprint. 
\end{itemize}

These conditions provide complementary trade-offs. Linear probing has the lowest training cost but cannot modify the representation itself. Full fine-tuning offers unrestricted adaptation, although its parameter, storage, and memory requirements are substantial and may be disadvantageous when several modality-specific backbones must be maintained. \gls{lora} greatly reduces the number of trainable parameters and permits compact task-specific adapters, but constrains the update to a selected set of low-rank transformations and may be sensitive to adapter placement and rank. \gls{qlora} further reduces the memory required by the frozen backbone, at the cost of introducing a quantized representation and additional implementation constraints.

For each dataset $\mathcal{D}$, modality $m$, and target $t$, the adaptation strategy--- full fine-tuning, \gls{lora}, or \gls{qlora}---achieving the highest validation macro F1 is selected as the adapted configuration transferred to Stage~II. Linear probing is retained as the non-adapted reference and does not participate in this selection. After selection, the classification head is discarded and the chosen adapted encoder $\mathcal{E}_{\phi}^{m}$ is fixed as a feature extractor; the original pretrained version of the same backbone, $\mathcal{E}_{\phi_0}^{m}$, provides the paired frozen condition used in Stage~II.

\subsection{Stage II: Multimodal Transfer Evaluation}
\label{sec:materials_methods:mlp}

The second stage investigates whether the target-specific gains obtained through unimodal adaptation are preserved when audio, image, and symbolic score representations are evaluated in a common multimodal setting. The selected adapted encoder for each modality is compared with the original frozen version of the same backbone, ensuring that the adaptation factor is evaluated through paired encoder identities rather than through independently selected frozen baselines.

We use the lightweight projection-based architecture described by Hidalgo-Centeno et al.~\cite{triscore} as a controlled downstream mechanism. A separate projection experiment is defined for each dataset and target label. Consequently, different supervision targets produce distinct task-specific projected spaces rather than a single representation shared across all targets. The projection architecture is not treated as an object of architectural comparison; its purpose is to test whether a common supervised mapping transforms frozen and adapted encoder outputs differently.

For every dataset--target experiment, the same projection function and parameters are applied to the normalized and dimension-matched representations from the three modalities. Projection training uses category-level supervision over the combined modality samples.

Let $\mathcal{E}_{\phi}^{m}$ denote the encoder--adaptation configuration selected in Stage~I for modality $m$. For input $x_i^m$, the adapted embedding is

\begin{equation}
z_i^m = \mathcal{E}_{\phi}^{m}(x_i^m).
\label{eq_adapted}
\end{equation}

Each embedding is $\ell_2$-normalized and, when necessary, zero-padded to the common input dimensionality of the projection, yielding $\bar{z}_i^m$. Thus, the complete collection of representations across modalities can be expressed as:

\begin{equation}
    \mathcal{Z} = \{ \bar{z}_i^m \;|\; m\in\mathcal{M} \}.
    \label{eq_representation_set}
\end{equation}

During projection training, a task-specific linear classifier is attached to $h_i$ and optimized using the combined training representations from the three modalities. The classifier supplies supervision to the projection model and is discarded after training. This yields the multimodal projector $f_{\phi_0}$ , through which, given a unimodal representation extracted from a pretrained encoder, a multimodal representation can be obtained:

\begin{equation}
h_i = f_{\theta}(\bar{z}_i^m),
\label{eq_projected}
\end{equation}
where $f_{\theta}$ is a shallow \gls{mlp} shared across all three modalities. Similar to Equation~\ref{eq_representation_set}, after training, the complete collection of projected embeddings can be obtained:

\begin{equation}
 \mathcal{H}=\{ f_{\theta}(\bar{z}_i^m) \;|\; m\in\mathcal{\mathcal{M}} \}.
\label{eq_projected_set}
\end{equation}

During projection training, a task-specific linear classifier is attached to each projected embedding $h_i^m$ and optimized using the corresponding labels $y_i^t$ across the combined embeddings from all three modalities. The classifier provides the supervision signal for learning the projection function and is discarded after training. The adapted encoders remain fixed throughout this stage.

\section{Experimental Setup}
\label{sec:experimental_setup}

This section instantiates the methodology introduced in Section~\ref{sec:materials_methods}. It first describes the two multimodal datasets, their target labels, and their data partitions. It then specifies the optimization and hyperparameter search used for unimodal adaptation, the selection of the encoders transferred to Stage~II, the training of the shared projection models, and the classification, retrieval, and neighborhood-analysis protocols. Validation data are used for all model and hyperparameter selection, while the held-out test partition is reserved for final evaluation.

\subsection{Corpora}
\label{sec:materials_methods:datasets}

We evaluate the methodology on two score-based multimodal datasets containing aligned audio, notation-level symbolic score, and sheet music image representations. The datasets differ in size, label granularity, modality content, and degree of cross-modal equivalence, providing complementary evaluation settings. Their main characteristics are summarized in Table~\ref{tab:dataset_summary}.

\subsubsection*{TriScore}
\label{sec:materials_methods:datasets:triscore}

\triscore{}~\cite{triscore} is a compact tri-modal dataset derived from MUSCUTS, which is itself based on the MUSCAT collection~\cite{muscat2024}. It contains short, tightly aligned excerpts of Western classical music. Metadata from the full MUSCAT corpus, specifically \emph{composer}, \emph{instrument}, and \emph{piece title}, is included to support classification and retrieval experiments.

Each aligned item consists of an audio excerpt, a symbolic score transcription in \KERN{} format, and a sheet music image rendered from the symbolic representation using the Verovio toolkit.\footnote{\url{https://verovio.humdrum.org}} The dataset contains 2,287 aligned triplets drawn from 111 musical works, 17 composers, and 8 instrument categories. Considering each modality independently, this corresponds to 6,861 modality-specific samples.

For the experiments in this work, we use the official training, validation, and test partitions, which follow an approximately 80/10/10 stratified split. The composer partition contains 1,829 training, 229 validation, and 229 test triplets. The class distributions are not uniform: composer classification is particularly imbalanced, while the instrument labels are dominated by piano and orchestral excerpts. The title distribution is comparatively less concentrated, although the number of fragments associated with each work remains variable.

In addition to exact cross-modal retrieval, where the objective is to recover the aligned counterpart of a query, \triscore{} supports \emph{piece-level retrieval}. Because several excerpts may originate from the same musical work, all fragments sharing the same piece-title label are considered relevant rather than only the exactly aligned triplet.

\subsubsection*{Musical Theme Dataset}
\label{sec:materials_methods:datasets:mtd}

The \gls{mtd}~\cite{ZalkowBAM20_MTD_TISMIR} is a collection of Western classical themes containing symbolic score encodings, sheet music images, and manually aligned audio excerpts. It comprises 2,068 aligned themes spanning 30 composers, 10 ensemble categories, and 13 instrument categories. Considering the three modalities independently, the dataset contains 6,204 modality-specific samples.

The original symbolic scores are distributed in MusicXML format. For this work, they were converted to \KERN{} using the humlib toolkit,\footnote{\url{https://humlib.humdrum.org}} providing a common symbolic representation across both datasets. The aligned themes are divided into training, validation, and test partitions using the same approximately 80/10/10 stratified protocol employed for \triscore{}. All three modalities associated with a theme are assigned to the same partition, preventing aligned counterparts from being distributed across different splits.

Unlike \triscore{}, \gls{mtd} consists of short monophonic themes. Its symbolic and visual modalities therefore encode mainly melodic information, whereas the corresponding audio excerpts may contain additional harmonic, timbral, and orchestral context. The modalities are consequently not always informationally equivalent, making \gls{mtd} a particularly challenging setting for multimodal representation learning.

\begin{table}[ht]
    \caption{Summary of the multimodal datasets used in this work. Sample counts refer to aligned triplets, each comprising audio, symbolic score, and sheet music image modalities.}
    \label{tab:dataset_summary}
    \centering
    \resizebox{.85\linewidth}{!}{%
    \begin{tabular}{lcc}
        \toprule
        \textbf{Property} & \textbf{\triscore{}} & \textbf{\gls{mtd}} \\
        \midrule
        Aligned triplets        & 2,287                         & 2,068 \\
        Composers               & 17                            & 30 \\
        Instrument categories   & 8                             & 13 \\
        Ensemble categories     & --                            & 10 \\
        Piece-level retrieval   & 111 works                     & -- \\
        Musical content         & Polyphonic excerpts           & Monophonic themes \\
        Symbolic representation & \KERN{}                       & MusicXML $\rightarrow$ \KERN{} \\
        Image source            & Rendered from symbolic score  & Original sheet music images \\
        Cross-modal equivalence & Closely matched excerpts      & Theme-level score, richer audio \\
        Data split              & Stratified 80/10/10           & Stratified 80/10/10 \\
        \bottomrule
    \end{tabular}%
    }
\end{table}

For \triscore{}, the unimodal supervision targets are \emph{instrument}, \emph{composer}, and \emph{title}. Instrument and composer are subsequently included in the factorial multimodal classification analysis. Title is used as a classification target during Stage~I so that encoder selection explicitly considers piece-related discrimination, and the final title-specific adapted-and-projected representation is evaluated through piece-level retrieval. For \gls{mtd}, the supervision targets are \emph{instrument}, \emph{composer}, and \emph{ensemble}; all three are included in the factorial multimodal classification analysis.

\subsection{Pretrained Unimodal Encoders}
\label{sec:materials_methods:encoders}

We evaluate nine publicly available pretrained transformer encoders, with three candidates assigned to each modality. The models were selected to represent contrasting pretraining domains and different degrees of compatibility with the corresponding musical input: music-specific, speech, and general-audio pretraining for audio; document, general-image, and vision--language pretraining for score images; and natural-language, source-code, and semantic-similarity pretraining for symbolic scores. This diversity allows Stage~I to test whether the effectiveness of full and parameter-efficient adaptation depends on the relationship between the pretraining domain and the target modality.

All checkpoints are obtained from the HuggingFace model repository.\footnote{\url{https://huggingface.co/models}} Table~\ref{tab:encoders} summarizes their pretraining domains and representation dimensionalities.

\subsubsection*{Audio encoders}

\mert{}~\cite{li2024mertacousticmusicunderstanding} (\href{https://huggingface.co/m-a-p/MERT-v1-330M}{\texttt{m-a-p/MERT-v1-330M}}) is pretrained specifically on large-scale music audio and is therefore the only audio encoder considered here whose pretraining objective is explicitly centered on music understanding.

Whisper~\cite{radford2022_whisper} (\href{https://huggingface.co/openai/whisper-medium}{\texttt{openai/whisper-medium}}) is a large-scale multilingual speech model. Only its encoder component is retained.

Wav2Vec2~\cite{quarta2024_wav2vec2_audioset} (\href{https://huggingface.co/ALM/wav2vec2-large-audioset}{\texttt{ALM/wav2vec2-large-audioset}}) is pretrained on AudioSet~\cite{gemmeke2017_audioset}, which contains speech, music, and environmental audio.

All three audio encoders produce 1,024-dimensional representations. Waveforms are resampled to 24~kHz for \mert{} and 16~kHz for Whisper and Wav2Vec2. Embeddings are obtained by mean-pooling the final hidden state over the temporal dimension.

\subsubsection*{Image encoders}

\dit{}~\cite{peebles2023_dit} (\href{https://huggingface.co/microsoft/dit-base-finetuned-rvlcdip}{\texttt{microsoft/dit-base-finetuned-rvlcdip}}) is pretrained for document image classification and is included to evaluate whether document-oriented representations transfer effectively to the structured layout of sheet music.

\vit{}~\cite{vit} (\href{https://huggingface.co/google/vit-base-patch16-224}{\texttt{google/vit-base-patch16-224}}) is a general-purpose vision transformer pretrained on natural images.

PaliGemma~\cite{steiner2024_paligemma2familyversatile} (\href{https://huggingface.co/google/paligemma2-3b-pt-896}{\texttt{google/paligemma2-3b-pt-896}}) is a vision--language model. Only its SigLIP~\cite{zhai2023sigmoidlosslanguageimage} vision tower and checkpoint-native visual projection are used; the language-model component is discarded.

\vit{} and \dit{} produce 768-dimensional representations. PaliGemma produces 4,096-dimensional vectors after projecting its 1,152-dimensional visual features. We treat the output of the checkpoint-native visual projection as the PaliGemma image representation. The CLS token is retained for \vit{}, while mean pooling over patch tokens is applied to \dit{} and PaliGemma.

\subsubsection*{Symbolic score encoders}

\roberta{}~\cite{liu2019robertarobustlyoptimizedbert} (\href{https://huggingface.co/FacebookAI/roberta-base}{\texttt{FacebookAI/roberta-base}}) is a general-purpose language model pretrained on natural-language corpora.

GraphCodeBERT~\cite{guo2021_graphcodebertpretrainingcoderepresentations} (\href{https://huggingface.co/microsoft/graphcodebert-base}{\texttt{microsoft/graphcodebert-base}}) is pretrained on source code and structural code information. It is included to evaluate whether code-oriented pretraining transfers to the structured syntax of \KERN{} notation.

SentenceTransformer~\cite{reimers2019_sentence} (\href{https://huggingface.co/sentence-transformers/all-mpnet-base-v2}{\texttt{sentence-transformers/all-mpnet-base-v2}}) is optimized for semantic similarity between sequences.

All symbolic encoders produce 768-dimensional representations. Scores are tokenized using the tokenizer associated with each model. Embeddings from \roberta{} and GraphCodeBERT are obtained by mean-pooling the final hidden state, whereas SentenceTransformer uses its standard attention-mask-weighted pooling operation.

\begin{table}[ht]
    \caption{Summary of the nine pretrained encoders evaluated in this work. Hidden size denotes the dimensionality of the extracted representation.}
    \label{tab:encoders}
    \centering
    \resizebox{\textwidth}{!}{%
    \begin{tabular}{llllc}
        \toprule
        \textbf{Modality} &
        \textbf{Encoder} &
        \textbf{HuggingFace model ID} &
        \textbf{Pretraining domain} &
        \textbf{Hidden size} \\
        \midrule

        \multirow{3}{*}{Audio}
        & \mert{}~\cite{li2024mertacousticmusicunderstanding}
        & \href{https://huggingface.co/m-a-p/MERT-v1-330M}
          {\texttt{m-a-p/MERT-v1-330M}}
        & Music audio & 1,024 \\

        & Whisper~\cite{radford2022_whisper}
        & \href{https://huggingface.co/openai/whisper-medium}
          {\texttt{openai/whisper-medium}}
        & Multilingual speech & 1,024 \\

        & Wav2Vec2~\cite{quarta2024_wav2vec2_audioset}
        & \href{https://huggingface.co/ALM/wav2vec2-large-audioset}
          {\texttt{ALM/wav2vec2-large-audioset}}
        & General audio & 1,024 \\

        \hdashline

        \multirow{3}{*}{Image}
        & \vit{}~\cite{vit}
        & \href{https://huggingface.co/google/vit-base-patch16-224}
          {\texttt{google/vit-base-patch16-224}}
        & General vision & 768 \\

        & \dit{}~\cite{peebles2023_dit}
        & \href{https://huggingface.co/microsoft/dit-base-finetuned-rvlcdip}
          {\texttt{microsoft/dit-base-finetuned-rvlcdip}}
        & Document images & 768 \\

        & PaliGemma~\cite{steiner2024_paligemma2familyversatile}
        & \href{https://huggingface.co/google/paligemma2-3b-pt-896}
          {\texttt{google/paligemma2-3b-pt-896}}
        & Vision--language & 4,096 \\

        \hdashline

        \multirow{3}{*}{Symbolic}
        & \roberta{}~\cite{liu2019robertarobustlyoptimizedbert}
        & \href{https://huggingface.co/FacebookAI/roberta-base}
          {\texttt{FacebookAI/roberta-base}}
        & General text & 768 \\

        & GraphCodeBERT~\cite{guo2021_graphcodebertpretrainingcoderepresentations}
        & \href{https://huggingface.co/microsoft/graphcodebert-base}
          {\texttt{microsoft/graphcodebert-base}}
        & Source code & 768 \\

        & SentenceTransformer~\cite{reimers2019_sentence}
        & \href{https://huggingface.co/sentence-transformers/all-mpnet-base-v2}
          {\texttt{sentence-transformers/all-mpnet-base-v2}}
        & Semantic similarity & 768 \\

        \bottomrule
    \end{tabular}%
    }
\end{table}

\subsection{Neural Configuration}
 
\subsubsection*{Unimodal Adaptation}
\label{sec:experimental_setup:unimodal}

All linear-probing and adaptation experiments are trained on the corresponding training partition using the cross-entropy objective. The encoder-specific input processing and pooling operations are described in Section~\ref{sec:materials_methods:encoders}. Model, checkpoint, and hyperparameter selection are based exclusively on validation macro F1.

All models are optimized with AdamW using an initial learning rate of $10^{-5}$, a weight decay of 0.01, and a batch size of 4. For the audio encoders---\mert{}, Whisper, and Wav2Vec2---the batch size is reduced to 1 because of memory constraints. Training uses early stopping based on validation macro F1 with a patience of 25 epochs. The checkpoint achieving the highest validation macro F1 is retained.

For \gls{lora} and \gls{qlora}, the low-rank updates defined in Equation~\ref{eq:lora_update} are inserted into the query, key, and value projections of each self-attention block. The corresponding module names are identified for each encoder architecture, including separate query, key, and value projections or a fused \texttt{qkv} projection where applicable. This procedure targets equivalent attention components across architectures. The task-specific classification head is trained jointly with the low-rank parameters.

The adapter rank $r$, scaling factor $\alpha$, and dropout probability are varied according to Table~\ref{tab:lora_hyperparameters}. The same search space is used for \gls{lora} and \gls{qlora}. No bias parameters are introduced or updated by the low-rank modules, and the adapter contribution is scaled by $\alpha/r$.

\begin{table}[ht]
    \caption{Hyperparameter configurations evaluated for \gls{lora} and \gls{qlora}. The same search space is used for both methods.}
    \label{tab:lora_hyperparameters}
    \centering
    \resizebox{.6\textwidth}{!}{%
    \begin{tabular}{cccc}
        \toprule
        \textbf{Configuration} &
        \textbf{Rank $r$} &
        \textbf{Scaling factor $\alpha$} &
        \textbf{Dropout} \\
        \midrule
        1 & 2 & 4 ($2r$)  & 0.1 \\
        2 & 2 & 8 ($4r$)  & 0.1 \\
        3 & 4 & 8 ($2r$)  & 0.1 \\
        4 & 4 & 8 ($2r$)  & 0.0 \\
        5 & 4 & 16 ($4r$) & 0.1 \\
        6 & 8 & 16 ($2r$) & 0.1 \\
        7 & 8 & 32 ($4r$) & 0.1 \\
        \bottomrule
    \end{tabular}
    }
\end{table}

For \gls{qlora}, the frozen backbone is loaded using 4-bit NF4 quantization with double quantization enabled. Computation is performed in \texttt{bfloat16} when supported by the hardware and in \texttt{float16} otherwise. Quantization reduces the memory required to store and process the frozen backbone, while the number of trainable parameters remains primarily determined by the low-rank configuration and classification head.

For each dataset--modality--target combination, the configuration transferred to Stage~II is selected among full fine-tuning, \gls{lora}, and \gls{qlora}. Frozen-encoder \gls{lp} is retained as a non-adapted reference and does not participate in the selection of the adapted condition. If two adapted configurations obtain the same validation score, the configuration with fewer trainable parameters is retained. The test partition is not used during model, checkpoint, or hyperparameter selection.

Once a configuration has been selected, the classification head is discarded. The selected encoder identity and adaptation parameters are then fixed for the subsequent multimodal evaluation.

\subsubsection*{Multimodal Projection}
\label{sec:experimental_setup:mlp}

For each dataset and target label, the adapted representations defined in Equation~\ref{eq_adapted} are extracted for the training, validation, and test partitions. All encoder parameters remain fixed during projection training. A projection model is trained independently for each dataset--target experiment, $f_{\theta}$, which receives representations from the pretrained encoder.

Each projection is trained using the corresponding training partition, and checkpoint selection is performed on validation data. No test samples are used for encoder selection, projection-model selection, or hyperparameter selection.

During projection training, samples from all three modalities contribute to a common task-specific classification objective. An auxiliary linear classification head is optimized jointly with the projection model and discarded after training. The resulting raw and projected representations are evaluated under the four conditions defined in Stage~II of Figure~\ref{fig:methodoloy}.

The projection dimensionality, optimization procedure, modality-balancing strategy, and classification objective follow the configuration reported in~\cite{triscore}. The same training protocol is applied to every dataset--target experiment and to both the frozen- and adapted-encoder projections.

\subsection{Evaluation}
\label{sec:experimental_setup:evaluation}

\subsubsection*{Factorial classification design}

Category-level multimodal classification is performed using \knn{} under all evaluations. The search space is formed from a \emph{combined} gallery containing samples from all three modalities. For every query, the $k$ nearest gallery samples are retrieved, and the predicted class is determined by majority vote. The value of $k$ is selected using the validation partition and subsequently fixed for test evaluation. Performance is then reported using macro F1.  

Equation~\ref{eq_adapted} defines the projection extraction through which the unimodal representations are projected to the multimodal space. To isolate the contribution of each stage, we follow a controlled $2\times2$ design. Let $\mathcal{V}$ denote the search embedding spaces and $\Phi$ the encoder parameterizations. The first factor indicates whether the encoder is \emph{frozen} ($\phi=\phi_0$) or \emph{adapted} ($\phi=\phi_{adapt}$), and the second whether the representations of the search space is before (\emph{Raw}, $v=\mathcal{Z}$),  or after (\emph{Projected}, $v=\mathcal{H}$) multimodal projection. For each $(v,\phi) \in \mathcal{V}\times\Phi$, this yields four scenarios in the evaluation:

\begin{equation}
    {(v,\phi)\in\mathcal{V}\times\Phi} 
    = \left\{ ({\mathcal{Z}},{\phi_0}),\;({\mathcal{H}},{\phi_0}),\; ({\mathcal{Z}},{\phi_{adapt}}),\; ({\mathcal{H}},{\phi_{adapt}}) \right\}.
    \label{eq_f1_family}
\end{equation}

This allows us to quantitatively measure the effects of the adaptation and projection separately. The raw conditions $(\phi_0, \mathcal{Z})$ and $(\phi_{adapt}, \mathcal{Z})$ are controls rather than intrinsically aligned multimodal spaces. Because the modality-specific encoders are trained independently, their coordinate systems are not expected to be semantically calibrated. Normalization and zero-padding permit a common downstream evaluation but do not establish correspondence between individual dimensions.

Thus, for each dataset and target label, we report the macro F1 metric over the \knn{} predictions for each of the aforementioned scenarios. In addition, to measure the effects of projection and adaptation, the following contrasts are obtained:

\begin{align}
    \Delta_{\mathrm{adapt}} = F1_{(v,{\phi_{adapt}})}-F1_{(v,{\phi_{0}})}, \quad
    \quad \Delta_{\mathrm{projection}} = F1_{({\mathcal{H}},{\phi})}-F1_{(\mathcal{Z},{\phi})}
\end{align}

$\Delta_{adapt}$ determines whether encoder adaptation improves multimodal classification without ($v=\mathcal{Z}$) or with shared projection ($v=\mathcal{H}$). $\Delta_{projection}$ determines whether projection improves the representations supplied by either frozen ($\phi =\phi_0$) or adapted ($\phi =\phi_{adapt}$) encoders.

A positive unimodal gain in the difference between frozen and adapted encoders is considered to provide evidence of transfer only when the adapted representation also improves over its paired frozen-encoder condition in the factorial evaluation. Thus, $\Delta_{\mathrm{adapt}}$ evaluates transfer before or after projection, whereas $\Delta_{\mathrm{adapt}}$ evaluates transfer after learning the shared task-specific mapping.

The complete factorial comparison is applied to composer and instrument classification on \triscore{}, and to composer, instrument, and ensemble classification on \gls{mtd}. Piece-level retrieval is evaluated separately using the final adapted-and-projected representation.

\subsubsection*{Piece-level retrieval}
\label{sec:experimental_setup:retrieval}

Piece-level retrieval is evaluated on \triscore{} using the title-specific adapted-and-projected representation, corresponding to condition $(\phi_{adapt}, \mathcal{H})$. The title-specific encoder and projection models are trained using title classification supervision, after which the resulting representation is evaluated through retrieval.

Given a query, gallery samples are ranked according to their distance from the query representation. A retrieval is considered correct when at least one of the first $N$ samples shares the same piece-title label as the query. Performance is reported using Top-$N$ accuracy following the protocol established for \triscore{}~\cite{triscore}.

Because several excerpts may originate from the same musical work, all gallery fragments sharing the query title are considered relevant rather than only the exactly aligned counterpart. The evaluation therefore measures whether excerpts belonging to the same work occupy nearby regions of the final title-specific representation space.

This experiment characterizes the practical piece-identification capability of the complete adapted-and-projected system. Because conditions $(v,\phi) \in \mathbf{}{V}\times\Phi$ are not evaluated under the same retrieval protocol, the resulting Top-$N$ scores are not used to determine whether retrieval performance originates from encoder adaptation, shared projection, or their interaction.

\section{Results}
\label{sec:results}

This section presents the results corresponding to the two central questions of the study. We first evaluate whether parameter-efficient adaptation provides a competitive alternative to full fine-tuning across datasets, modalities, and supervision targets. Section~\ref{sec:results:unimodal_ft} reports the complete unimodal validation results, while Section~\ref{sec:results:full_lora_qlora} compares adaptation strategies in terms of predictive performance, trainable-parameter requirements, and the configurations transferred to the multimodal stage.

We then use the held-out test partition to address the second question: whether the improvements obtained through the selected unimodal adaptation strategies translate into better multimodal classification. Section~\ref{sec:results:final_impact} analyzes the complete $2\times2$ design crossing encoder adaptation with task-specific shared projection. Piece-level retrieval is evaluated separately using the final adapted-and-projected title representation. Finally, Section~\ref{sec:results:analysis} examines cross-modal exclusion and local neighborhood composition to determine whether category-level organization is accompanied by cross-modal mixing at the instance level.

\subsection{Unimodal adaptation results}
\label{sec:results:unimodal_ft}

We first examine the complete set of unimodal adaptation experiments before selecting the encoder configurations transferred to the multimodal stage. All results in this subsection correspond to the validation partition and are used exclusively for model selection. The held-out test partition is reserved for the multimodal evaluation.

\begin{figure*}[ht]
    \centering
    \includegraphics[width=\textwidth]{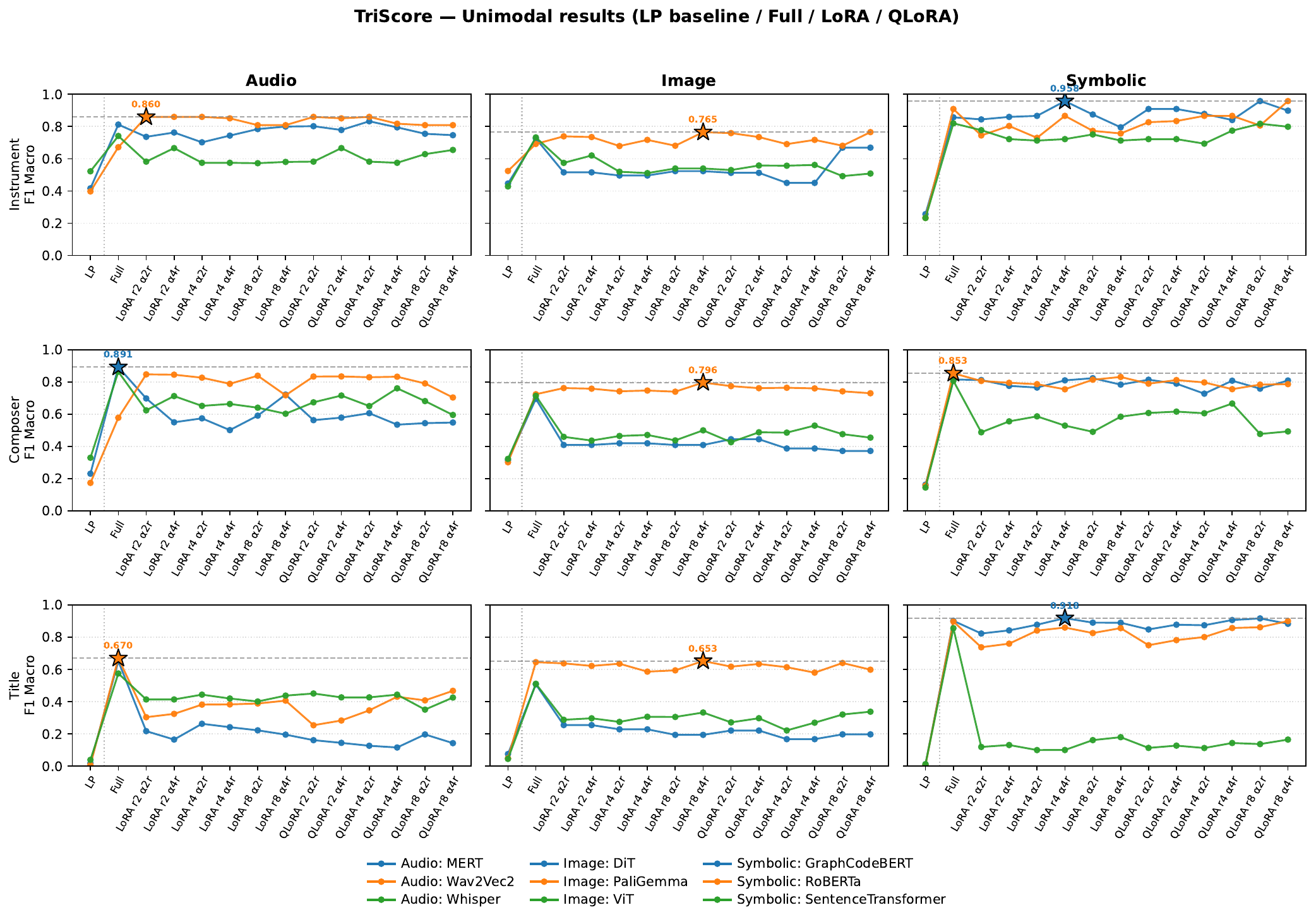}
    \caption{Validation macro F1 for all unimodal adaptation configurations on \triscore{}. These scores are used for encoder selection and do not correspond to the final multimodal test evaluation. Each row corresponds to a target label and each column to a modality. Lines represent the three encoders evaluated for each modality, starting from the frozen-encoder \gls{lp} baseline and followed by the full fine-tuning and \gls{lora}/\gls{qlora} configurations. Stars indicate the best encoder--adaptation configuration within each panel.}
    \label{fig:unimodal_triscore}
\end{figure*}

\begin{figure*}[ht]
    \centering
    \includegraphics[width=\textwidth]{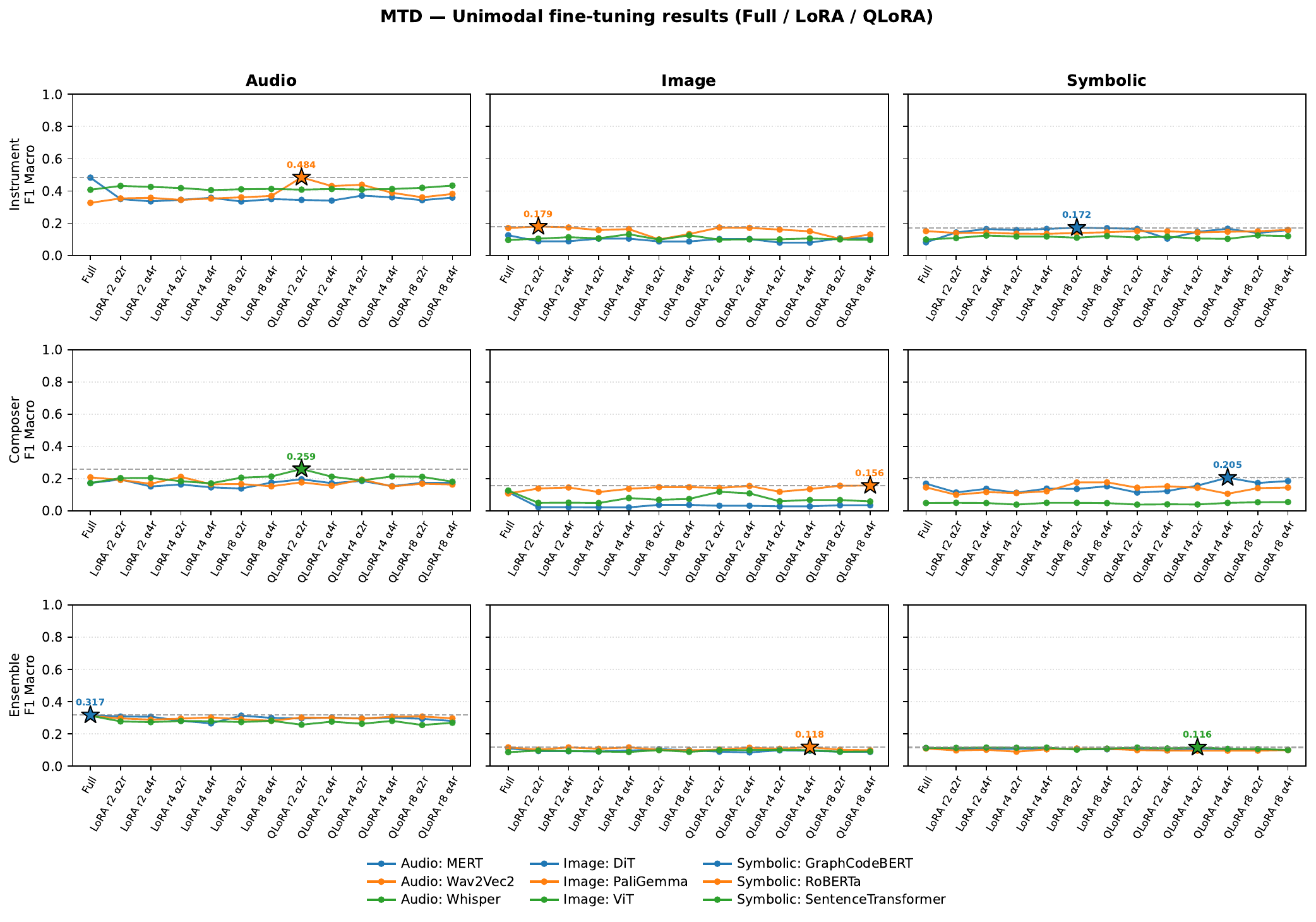}
    \caption{Validation macro F1 for all unimodal adaptation configurations on \gls{mtd}. These scores are used for encoder selection and do not correspond to the final multimodal test evaluation. Each row corresponds to a target label and each column to a modality. Lines represent the three encoders evaluated for each modality, starting from the frozen-encoder \gls{lp} baseline and followed by the full fine-tuning and \gls{lora}/\gls{qlora} configurations. Stars indicate the best encoder--adaptation configuration within each panel.}
    \label{fig:unimodal_mtd}
\end{figure*}

Figures~\ref{fig:unimodal_triscore} and~\ref{fig:unimodal_mtd} report the validation macro F1 obtained by every encoder under full fine-tuning and the evaluated \gls{lora} and \gls{qlora} configurations. Each panel also includes the \gls{lp} result obtained from the same frozen pretrained encoder, dataset, and target label.

The results are organized by target label and modality. This representation allows us to examine the best-performing configuration, the stability of each encoder across parameter-efficient settings, and the extent to which task-specific adaptation improves over the discriminative information already present in the frozen representation.

For \triscore{}, macro F1 is computed for \emph{instrument}, \emph{composer}, and \emph{title}. Although title is evaluated through piece-level retrieval in the final multimodal experiment, it is formulated as a classification target at this stage so that encoder selection explicitly considers piece-related discrimination. Instrument and composer are subsequently included in the factorial multimodal classification analysis, whereas title is used to evaluate the final adapted-and-projected retrieval representation.

\subsubsection{Overall comparison across adaptation configurations}
\label{sec:results:unimodal_overview}

The two datasets exhibit markedly different unimodal behavior. On \triscore{}, the results reveal substantial differences across modalities and pretrained backbones. Audio and symbolic encoders generally achieve the strongest validation scores, while image performance depends more strongly on the selected encoder. The panel-level maxima for audio are 0.860, 0.891, and 0.670 for instrument, composer, and title, respectively. The corresponding maxima are 0.765, 0.796, and 0.653 for image, and 0.958, 0.853, and 0.919 for symbolic scores.

For every panel, the best configuration found under each adaptation strategy improves over the corresponding encoder-specific \gls{lp} result, although individual \gls{lora} and \gls{qlora} configurations may perform below the frozen reference. The magnitude of the improvement is strongly target-dependent. Gains for instrument and composer generally range between approximately 20 and 70~\gls{pp}. Title produces particularly large gains because several \gls{lp} results are close to zero: adaptation increases macro F1 by more than 60~\gls{pp} for every audio encoder and by as much as 90.8~\gls{pp} for GraphCodeBERT on symbolic scores.

Within the audio modality, Wav2Vec2 provides the highest instrument and title scores, whereas \mert{} obtains the best composer result. Whisper remains competitive for composer classification, reaching 0.863, but does not produce a panel-level maximum. In the image modality, PaliGemma consistently outperforms \dit{} and \vit{}, with its largest advantage appearing for title classification. For symbolic scores, GraphCodeBERT obtains the strongest instrument and title results, while \roberta{} leads on composer. SentenceTransformer is generally weaker than the other symbolic encoders, although it remains comparatively competitive under full fine-tuning for composer and title.

The most effective adaptation strategy on \triscore{} also varies across encoders and targets. Full fine-tuning is selected for audio composer and title and for symbolic composer. \gls{lora} produces the selected configurations for audio instrument, all three image targets, and symbolic instrument and title. \gls{qlora} remains competitive, including a macro F1 of 0.958 with \roberta{} for symbolic instrument, but does not provide a unique selected maximum after applying the parameter-efficiency tie-breaking criterion.

On \gls{mtd}, absolute performance is substantially lower across all modalities and labels. Audio remains the strongest modality, with panel-level maxima of 0.484 for instrument, 0.259 for composer, and 0.317 for ensemble. Image results remain below 0.18, while symbolic results peak at 0.205.

Despite these lower absolute values, the best adaptation configuration improves over the corresponding encoder-specific \gls{lp} result in every panel. The gains generally range from approximately 1 to 24~\gls{pp}, which is considerably smaller than on \triscore{} and reflects the more limited target-discriminative information available in several \gls{mtd} modality--label combinations.

The modality-specific pattern is also more asymmetric on \gls{mtd}. Audio encoders retain useful information, particularly for instrument and ensemble, whereas image and symbolic encoders remain near the lower end of the performance range. PaliGemma nevertheless provides the strongest image result for all three targets, and GraphCodeBERT leads the symbolic instrument and composer tasks. For symbolic ensemble classification, differences among the three encoders are small, with SentenceTransformer obtaining the highest score by a narrow margin.

Across both datasets, changing the pretrained backbone generally has a larger effect than modifying the adapter rank, scaling factor, or dropout probability. Several \gls{lora} and \gls{qlora} curves remain relatively stable across hyperparameter configurations, particularly on \gls{mtd}. This stability is not universal, however. For example, SentenceTransformer degrades substantially under several parameter-efficient configurations on the \triscore{} composer and title tasks relative to full fine-tuning. The \gls{lp} result itself remains constant for each encoder--dataset--target combination because it depends only on the original frozen pretrained representation.

\subsubsection{Best encoder--adaptation configurations}
\label{sec:results:unimodal_best}

Tables~\ref{tab:unimodal_results_triscore} and~\ref{tab:unimodal_results_mtd} summarize the best validation configuration obtained for each encoder and target label. Whereas Figures~\ref{fig:unimodal_triscore} and~\ref{fig:unimodal_mtd} show the complete experimental search, the tables retain only the highest-scoring adaptation configuration for each encoder.

Boldface indicates the encoder--adaptation pair selected for the corresponding dataset, modality, and target label. Alongside each macro F1 value, we report the absolute gain over the encoder- and target-specific \gls{lp} result in \gls{pp}. Absolute differences are used instead of relative percentages because several frozen-encoder results, particularly for title, are close to zero and would produce inflated relative gains with limited interpretive value.

\begin{table}[H]
    \caption{Best validation configuration per encoder and target label on \triscore{}. \emph{Method} indicates the selected adaptation strategy (full fine-tuning, \gls{lora}, or \gls{qlora}); $r$ and $\alpha$ report the adapter rank and scaling factor and are not applicable to full fine-tuning; \emph{Trainable Params} reports the number of optimized parameters; and \emph{Macro F1} reports validation performance, with the gain over the linear-probing baseline shown in parentheses (in \gls{pp}). The best encoder--adaptation configuration for each modality and target is highlighted in bold.}
    \label{tab:unimodal_results_triscore}
    \centering
    \resizebox{.9\textwidth}{!}{%
    \begin{tabular}{lllccccc}
    \toprule
    \textbf{Modality} &
    \textbf{Label} &
    \textbf{Encoder} &
    \textbf{Method} &
    \boldmath$r$ &
    \boldmath$\alpha$ &
    \textbf{Trainable Params} &
    \textbf{Macro F1} \\
    \midrule

    \multirow{9}{*}{Audio}
        & \multirow{3}{*}{Instrument}
            & \mert{}              & \gls{qlora} & 4  & $2r$ & 1.6M & 0.832 \scriptsize(\up{41.6}) \\
        &   & Whisper              & Full        & -- & --   & 306M & 0.741 \scriptsize(\up{21.9}) \\
        &   & \textbf{Wav2Vec2}    & \gls{lora}  & 2  & $2r$ & 1.4M & \textbf{0.860} \scriptsize(\up{46.3}) \\
    \cdashline{2-8}

        & \multirow{3}{*}{Composer}
            & \textbf{\mert{}}     & Full        & -- & --   & 316M & \textbf{0.891} \scriptsize(\up{66.1}) \\
        &   & Whisper              & Full        & -- & --   & 306M & 0.863 \scriptsize(\up{53.4}) \\
        &   & Wav2Vec2             & \gls{lora}  & 2  & $2r$ & 1.4M & 0.847 \scriptsize(\up{67.4}) \\
    \cdashline{2-8}

        & \multirow{3}{*}{Title}
            & \mert{}              & Full & -- & -- & 316M & 0.652 \scriptsize(\up{63.4}) \\
        &   & Whisper              & Full & -- & -- & 306M & 0.575 \scriptsize(\up{53.6}) \\
        &   & \textbf{Wav2Vec2}    & Full & -- & -- & 316M & \textbf{0.670} \scriptsize(\up{66.3}) \\

    \midrule

    \multirow{9}{*}{Image}
        & \multirow{3}{*}{Instrument}
            & \dit{}               & Full        & -- & --   & 86.4M & 0.725 \scriptsize(\up{27.9}) \\
        &   & \vit{}               & Full        & -- & --   & 87.0M & 0.732 \scriptsize(\up{30.4}) \\
        &   & \textbf{PaliGemma}   & \gls{lora}  & 8  & $4r$ & 21.9M & \textbf{0.765} \scriptsize(\up{24.1}) \\
    \cdashline{2-8}

        & \multirow{3}{*}{Composer}
            & \dit{}               & Full        & -- & --   & 86.4M & 0.695 \scriptsize(\up{38.1}) \\
        &   & \vit{}               & Full        & -- & --   & 87.0M & 0.714 \scriptsize(\up{39.2}) \\
        &   & \textbf{PaliGemma}   & \gls{lora}  & 8  & $4r$ & 21.9M & \textbf{0.796} \scriptsize(\up{49.5}) \\
    \cdashline{2-8}

        & \multirow{3}{*}{Title}
            & \dit{}               & Full        & -- & --   & 86.4M & 0.508 \scriptsize(\up{43.3}) \\
        &   & \vit{}               & Full        & -- & --   & 87.0M & 0.511 \scriptsize(\up{46.6}) \\
        &   & \textbf{PaliGemma}   & \gls{lora}  & 8  & $4r$ & 21.9M & \textbf{0.653} \scriptsize(\up{60.4}) \\

    \midrule

    \multirow{9}{*}{Symbolic}
        & \multirow{3}{*}{Instrument}
            & \roberta{}                  & \gls{qlora} & 8  & $4r$ & 1.0M & 0.958 \scriptsize(\up{72.5}) \\
        &   & \textbf{GraphCodeBERT}      & \gls{lora}  & 4  & $4r$ & 724K & \textbf{0.958} \scriptsize(\up{70.2}) \\
        &   & SentenceTransformer         & Full        & -- & --   & 110M & 0.820 \scriptsize(\up{58.8}) \\
    \cdashline{2-8}

        & \multirow{3}{*}{Composer}
            & \textbf{\roberta{}}         & Full        & -- & --   & 125M & \textbf{0.853} \scriptsize(\up{69.7}) \\
        &   & GraphCodeBERT               & \gls{lora}  & 8  & $2r$ & 724K & 0.823 \scriptsize(\up{66.1}) \\
        &   & SentenceTransformer         & Full        & -- & --   & 110M & 0.807 \scriptsize(\up{66.3}) \\
    \cdashline{2-8}

        & \multirow{3}{*}{Title}
            & \roberta{}                  & \gls{qlora} & 8  & $4r$ & 1.0M & 0.901 \scriptsize(\up{89.4}) \\
        &   & \textbf{GraphCodeBERT}      & \gls{lora}  & 4  & $4r$ & 724K & \textbf{0.919} \scriptsize(\up{90.8}) \\
        &   & SentenceTransformer         & Full        & -- & --   & 110M & 0.856 \scriptsize(\up{84.4}) \\

    \bottomrule
    \end{tabular}%
    }
\end{table}

\begin{table}[H]
    \caption{Best validation configuration per encoder and target label on \gls{mtd}. \emph{Method} indicates the selected adaptation strategy (full fine-tuning, \gls{lora}, or \gls{qlora}); $r$ and $\alpha$ report the adapter rank and scaling factor and are not applicable to full fine-tuning; \emph{Trainable Params} reports the number of optimized parameters; and \emph{Macro F1} reports validation performance, with the gain over the linear-probing baseline shown in parentheses (in \gls{pp}). The best encoder--adaptation configuration for each modality and target is highlighted in bold. In the case of tied validation macro F1, the configuration with fewer trainable parameters is selected.}
    \label{tab:unimodal_results_mtd}
    \centering
    \resizebox{.9\textwidth}{!}{%
    \begin{tabular}{lllccccc}
    \toprule
    \textbf{Modality} &
    \textbf{Label} &
    \textbf{Encoder} &
    \textbf{Method} &
    \boldmath$r$ &
    \boldmath$\alpha$ &
    \textbf{Trainable Params} &
    \textbf{Macro F1} \\
    \midrule

    \multirow{9}{*}{Audio}
        & \multirow{3}{*}{Instrument}
            & \mert{}                 & Full        & -- & --   & 316M & 0.482 \scriptsize(\up{23.6}) \\
        &   & Whisper                 & \gls{qlora} & 8  & $4r$ & 1.4M & 0.433 \scriptsize(\up{19.3}) \\
        &   & \textbf{Wav2Vec2}       & \gls{qlora} & 2  & $2r$ & 1.4M & \textbf{0.484} \scriptsize(\up{23.0}) \\
    \cdashline{2-8}

        & \multirow{3}{*}{Composer}
            & \mert{}                 & Full        & -- & --   & 316M & 0.172 \scriptsize(\up{8.2}) \\
        &   & \textbf{Whisper}        & \gls{qlora} & 2  & $2r$ & 1.4M & \textbf{0.259} \scriptsize(\up{19.5}) \\
        &   & Wav2Vec2                & \gls{lora}  & 4  & $2r$ & 1.4M & 0.221 \scriptsize(\up{15.6}) \\
    \cdashline{2-8}

        & \multirow{3}{*}{Ensemble}
            & \textbf{\mert{}}        & Full        & -- & --   & 316M & \textbf{0.317} \scriptsize(\up{9.9}) \\
        &   & Whisper                 & Full        & -- & --   & 306M & 0.313 \scriptsize(\up{13.7}) \\
        &   & Wav2Vec2                & \gls{lora}  & 4  & $2r$ & 1.4M & 0.315 \scriptsize(\up{11.9}) \\

    \midrule

    \multirow{9}{*}{Image}
        & \multirow{3}{*}{Instrument}
            & \dit{}                  & Full        & -- & --   & 86.4M & 0.125 \scriptsize(\up{3.0}) \\
        &   & \vit{}                  & \gls{lora}  & 4  & $4r$ & 707K & 0.131 \scriptsize(\up{5.3}) \\
        &   & \textbf{PaliGemma}      & \gls{lora}  & 2  & $2r$ & 21.9M & \textbf{0.179} \scriptsize(\up{10.8}) \\
    \cdashline{2-8}

        & \multirow{3}{*}{Composer}
            & \dit{}                  & Full        & -- & --   & 86.4M & 0.116 \scriptsize(\up{7.7}) \\
        &   & \vit{}                  & Full        & -- & --   & 87.0M & 0.125 \scriptsize(\up{10.5}) \\
        &   & \textbf{PaliGemma}      & \gls{qlora} & 8  & $4r$ & 21.9M & \textbf{0.156} \scriptsize(\up{14.1}) \\
    \cdashline{2-8}

        & \multirow{3}{*}{Ensemble}
            & \dit{}                  & Full        & -- & --   & 86.4M & 0.111 \scriptsize(\up{0.8}) \\
        &   & \vit{}                  & \gls{qlora} & 4  & $2r$ & 707K & 0.102 \scriptsize(\up{1.5}) \\
        &   & \textbf{PaliGemma}      & \gls{qlora} & 4  & $4r$ & 21.9M & \textbf{0.118} \scriptsize(\up{4.8}) \\

    \midrule

    \multirow{9}{*}{Symbolic}
        & \multirow{3}{*}{Instrument}
            & \roberta{}                  & \gls{qlora} & 8  & $4r$ & 1.0M & 0.158 \scriptsize(\up{7.7}) \\
        &   & \textbf{GraphCodeBERT}      & \gls{lora}  & 8  & $2r$ & 724K & \textbf{0.172} \scriptsize(\up{9.5}) \\
        &   & SentenceTransformer         & \gls{qlora} & 8  & $2r$ & 724K & 0.124 \scriptsize(\up{6.5}) \\
    \cdashline{2-8}

        & \multirow{3}{*}{Composer}
            & \roberta{}                  & \gls{lora}  & 8  & $4r$ & 1.0M & 0.177 \scriptsize(\up{16.2}) \\
        &   & \textbf{GraphCodeBERT}      & \gls{qlora} & 4  & $4r$ & 724K & \textbf{0.205} \scriptsize(\up{19.0}) \\
        &   & SentenceTransformer         & \gls{qlora} & 8  & $4r$ & 724K & 0.054 \scriptsize(\up{3.9}) \\
    \cdashline{2-8}

        & \multirow{3}{*}{Ensemble}
            & \roberta{}                      & Full        & -- & --   & 125M & 0.110 \scriptsize(\up{5.0}) \\
        &   & GraphCodeBERT                   & Full        & -- & --   & 125M & 0.112 \scriptsize(\up{2.6}) \\
        &   & \textbf{SentenceTransformer}    & \gls{qlora} & 4  & $2r$ & 724K & \textbf{0.116} \scriptsize(\up{5.5}) \\

    \bottomrule
    \end{tabular}%
    }
\end{table}

When two configurations obtain the same validation macro F1, the configuration with fewer trainable parameters is selected. This criterion applies to the \triscore{} symbolic instrument task, where GraphCodeBERT with \gls{lora} and \roberta{} with \gls{qlora} both reach 0.958. GraphCodeBERT is retained because it optimizes 724K parameters instead of 1.0M. Both configurations improve substantially over their respective \gls{lp} results (\up{70.2} and \up{72.5}, respectively), showing that the tie occurs between two effective adaptation configurations rather than between representations that remain close to their frozen baselines.

\subsection{Parameter-efficient adaptation versus full fine-tuning}
\label{sec:results:full_lora_qlora}

Parameter-efficient adaptation is selected in 14 of the 18 dataset--modality--target combinations. Of these, eight use \gls{lora} and six use \gls{qlora}, while full fine-tuning is selected in the remaining four combinations. This overall distribution shows that \gls{peft} provides the strongest validation configuration in most experimental conditions, although its relative effectiveness remains dependent on the pretrained backbone, dataset, modality, and supervision target.

On \triscore{}, six of the nine selected configurations use \gls{lora}. These correspond to audio instrument, all three image targets, and symbolic instrument and title. Full fine-tuning is selected for audio composer and title and for symbolic composer. No \gls{qlora} configuration is uniquely selected, although \roberta{} with \gls{qlora} ties the selected GraphCodeBERT \gls{lora} configuration at 0.958 macro F1 for symbolic instrument. GraphCodeBERT is retained because it optimizes fewer trainable parameters.

On \gls{mtd}, \gls{peft} is selected in eight of the nine modality--target combinations. \gls{qlora} is selected for audio instrument and composer, image composer and ensemble, and symbolic composer and ensemble. \gls{lora} is selected for image instrument and symbolic instrument. Full fine-tuning is selected only for audio ensemble. The prevalence of \gls{peft} on this dataset should nevertheless be interpreted alongside the low absolute macro F1 values, particularly for image and symbolic representations, where differences among adaptation strategies are often small.

The reductions in trainable parameters are substantial. Selected \gls{peft} configurations optimize approximately 1.4--1.6M parameters for audio encoders, compared with more than 300M under full fine-tuning. For \dit{} and \vit{}, fewer than one million parameters are optimized instead of approximately 86--87M. The selected symbolic configurations require approximately 724K--1.0M parameters, compared with 110--125M under full fine-tuning. PaliGemma requires 21.9M trainable parameters under the selected \gls{peft} configurations. Thus, \gls{peft} often reduces the number of optimized parameters by approximately two orders of magnitude while retaining competitive predictive performance.

Overall, \gls{lora} and \gls{qlora} frequently match or outperform full fine-tuning while optimizing substantially fewer parameters. They are not uniformly superior, however: full fine-tuning remains preferable for several \triscore{} audio and symbolic targets and for the \gls{mtd} audio ensemble. Moreover, changing the pretrained backbone often has a larger effect on validation macro F1 than choosing between \gls{lora} and \gls{qlora} or modifying their rank, scaling factor, and dropout settings.

\subsection{Transfer from unimodal adaptation to multimodal classification}
\label{sec:results:final_impact}

The preceding analyses address the first central question by showing that parameter-efficient methods provide 14 of the 18 selected modality--target configurations and frequently match or outperform full fine-tuning with substantially fewer trainable parameters. We now address the second question: whether the improvements obtained through these target-specific selections translate into better multimodal classification.

The adapted conditions in this analysis use the strategy selected independently for each modality in Stage~I (Sec.~\ref{sec:materials_methods:ft}). Consequently, a multimodal encoder triplet may consist entirely of parameter-efficiently adapted encoders or may combine \gls{lora}, \gls{qlora}, and full fine-tuning across modalities. The factorial analysis therefore evaluates the transfer of the selected adaptation pipeline as a whole; it does not isolate the separate multimodal effects of the three adaptation strategies.

Table~\ref{tab:ablation_full} reports test macro F1 using a combined gallery for all four conditions. It also reports the effect of adaptation without and with projection, $\Delta_{adapt}$, and the effect of projection without and with adaptation, $\Delta_{projection}$.

\begin{table}[H]
    \caption{Controlled $2\times2$ factorial evaluation crossing encoder adaptation (frozen vs.\ adapted) with shared projection (raw vs.\ projected). Scores are macro F1 values in \gls{pp}, combined-gallery average. $\Delta_{adapt}$ reports the effect of adaptation without and with projection, while $\Delta_{projection}$ reports the effect of projection without and with adaptation.}
    \label{tab:ablation_full}
    \centering
    \resizebox{\textwidth}{!}{%
    \begin{tabular}{llccc:ccc:cc}
    \toprule
    & & \multicolumn{3}{c:}{Frozen ($\phi_0$)} & \multicolumn{3}{c:}{Adapted ($\phi_{adapt}$)} & \multicolumn{2}{c}{$\Delta_{adapt}$} \\ 
    \cmidrule(lr){3-5} \cmidrule(lr){6-8} \cmidrule(lr){9-10}
    \textbf{Dataset} & \textbf{Label} &
    Raw ($\mathcal{Z}$) & Projected ($\mathcal{H}$)  & $\Delta_{projection}$ & Raw ($\mathcal{Z}$) & Projected ($\mathcal{H}$)  & $\Delta_{projection}$ & Raw ($\mathcal{Z}$) & Projected ($\mathcal{H}$)  \\
    \midrule
    \multirow{2}{*}{\triscore{}}
        & Composer   & 42.09 & 50.63 & \up{8.54}  & 61.55 & 74.99 & \up{13.44} & \up{19.46} & \up{24.36} \\
        & Instrument & 59.72 & 74.39 & \up{14.67} & 69.26 & 70.03 & \up{0.77}  & \up{9.54}  & \down{4.36} \\
    \midrule
    \multirow{3}{*}{\gls{mtd}}
        & Instrument & 15.38 & 17.28 & \up{1.90}  & 16.15 & 16.60 & \up{0.45}  & \up{0.77}  & \down{0.68} \\
        & Ensemble   & 11.91 & 16.24 & \up{4.33}  & 16.64 & 16.29 & \down{0.35} & \up{4.73}  & \up{0.05}  \\
        & Composer   & 4.38  & 7.34  & \up{2.96}  & 11.41 & 8.58  & \down{2.83} & \up{7.03}  & \up{1.24}  \\
    \bottomrule
    \end{tabular}%
    }
\end{table}

The factorial results show that improvements in unimodal discrimination do not translate uniformly into improvements in multimodal classification. The relationship between adaptation and projection depends strongly on the dataset and target label. Three broad interaction patterns can be identified.

\subsubsection*{Complementary effects}

The clearest positive transfer occurs for \triscore{} composer. Adaptation improves the raw representation from 42.09 to 61.55 macro F1 ($\Delta_{adapt}=19.46$~pp), and its benefit becomes larger when the shared projection is present ($\Delta_{adapt}=24.36$~pp). Projection is also more effective on the adapted representation: its $\Delta_{projection}$ increases from 8.54~pp on frozen features to 13.44~pp on adapted features.

The complete system reaches 74.99 macro F1, an improvement of 32.90~pp over the frozen--raw condition. Composer on \triscore{} is therefore the only target for which adaptation and projection clearly reinforce one another in both directions. In this setting, the information emphasized through unimodal adaptation appears to remain compatible with, and to benefit from, the subsequent shared mapping.

\subsubsection*{Redundant effects}

For \triscore{} instrument, adaptation and projection are individually beneficial but do not combine additively. Adaptation alone improves macro F1 by 9.55~pp, while projection alone improves it by 14.67~pp. Once the projection is applied, however, replacing the frozen encoders with their adapted counterparts reduces performance by 4.36~pp. Conversely, applying projection to the adapted representation provides only a 0.77~pp gain. The two transformations therefore appear to capture overlapping task-relevant structure, with the frozen--projected condition $(\phi_0, \mathcal{H})$ outperforming the complete condition  $(\phi_{adapt}, \mathcal{H})$.

A similar but less pronounced pattern occurs for the \gls{mtd} instrument. Projection alone provides a 1.91~pp gain and adaptation alone a 0.77~pp gain. Combining both produces only 16.60 macro F1: adaptation after projection decreases performance by 0.68~pp, while projection after adaptation contributes only 0.46~pp.

For \gls{mtd} ensemble, adaptation and projection produce similar standalone improvements over $(\phi_o, \mathcal{Z})$: 4.74 and 4.34~pp, respectively. Their combination contributes essentially no additional gain. Adaptation changes $(\phi_0, \mathcal{H})$ by only $+0.05$~pp, and projection changes $(\phi_{adapt}, \mathcal{Z})$ by $-0.35$~pp. These results indicate that the two mechanisms capture largely redundant information for this target.

\subsubsection*{Interfering effects}

The strongest interference occurs for the \gls{mtd} composer. Adaptation substantially improves the raw representation, increasing macro F1 from 4.38 to 11.41 ($F1_{(\mathcal{Z},\phi_{adapt})}-F1_{(\mathcal{Z},\phi_0)}=7.03$~pp). Projection alone also improves the frozen features by 2.96~pp. However, applying the projection to the adapted representation reduces performance by 2.83~pp.

Adaptation remains mildly beneficial relative to the frozen--projected condition ($\Delta_{adapt}=1.24$~pp), but most of the gain observed before projection is lost. The adapted representation is therefore more discriminative in the raw evaluation yet is not organized in a way that the category-supervised shared projection can preserve effectively.

A small negative projection effect is also observed for the \gls{mtd} ensemble ($\Delta_{projection}=-0.35$~pp), but the composer result is substantially larger and occurs in a lower-performing representation space. It constitutes the clearest case in which improved unimodal discrimination fails to survive the multimodal projection stage.

\subsubsection*{Overall transfer pattern}

Taken together, the results show that encoder adaptation is not an inherently transferable improvement. It can provide a better basis for multimodal classification, as in \triscore{} composer; become largely redundant once a shared projection is learned, as in \triscore{} instrument and the \gls{mtd} instrument and ensemble tasks; or lose part of its benefit after projection, as in \gls{mtd} composer.

The contrast between the datasets is consistent with their structural differences. On \triscore{}, the three modalities represent closely corresponding excerpts, allowing the composer-related information emphasized by the adapted encoders to be integrated successfully. On \gls{mtd}, the visual and symbolic inputs describe short monophonic themes, whereas the audio may contain additional harmonic, timbral, orchestral, and contextual information. Under this asymmetry, adaptation may emphasize modality-specific predictive cues that are unavailable, or represented differently, in the other modalities. A single shared projection can therefore have difficulty preserving those gains consistently.

\subsection{Qualitative visualization of the factorial representations}
\label{sec:results:tsne}

To illustrate the factorial patterns identified in Section~\ref{sec:results:final_impact}, Figures~\ref{fig:tsne_triscore_composer}, \ref{fig:tsne_triscore_instrument}, and \ref{fig:tsne_mtd_composer} show t-SNE projections of representative cases for \triscore{} composer, \triscore{} instrument, and \gls{mtd} composer, respectively. Colors denote class and marker shapes denote modality. Each panel is obtained through an independent PCA+t-SNE fit and should only be interpreted qualitatively, comparing the relative organization of color and shape within a panel rather than coordinates across panels.

\subsubsection*{\triscore{} composer: a complementary case}

\begin{figure}[ht!]
    \centering
    \includegraphics[width=\textwidth]{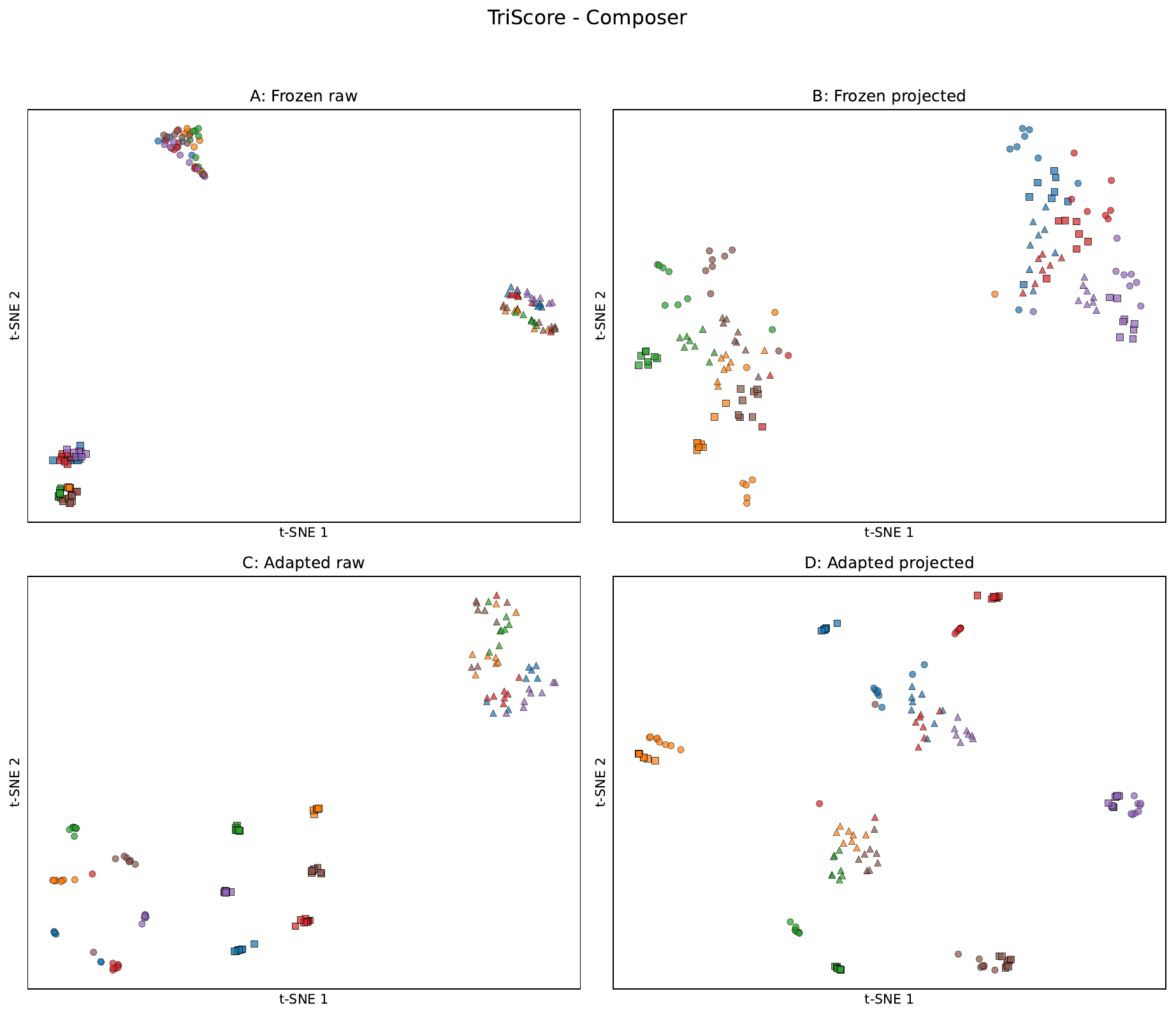}
    \caption{t-SNE visualization of the four factorial conditions for \triscore{} composer: $\{F1_e^\phi\}_{(e,\phi)\in\mathcal{S}\times\Phi} 
    = \left\{ F1_{z}^{\phi_0},\;F1_{h}^{\phi_0},\; F1_{z}^{\phi_{adapt}},\; F1_{h}^{\phi_{adapt}} \right\}.$. Colors indicate composer class; marker shapes indicate modality.}
    \label{fig:tsne_triscore_composer}
\end{figure}

Figure~\ref{fig:tsne_triscore_composer} shows all four factorial conditions for \triscore{} composer, the case with the clearest complementary interaction in Table~\ref{tab:ablation_full}. In $(\phi_0, \mathcal{Z})$(frozen--raw), the three modalities occupy essentially disjoint regions of the plot, and class colors are thoroughly mixed within each region: this is expected, since $(\phi_0, \mathcal{Z})$and  $(\phi_{adapt}, \mathcal{Z})$are controls whose per-modality coordinate systems are not calibrated against one another, and normalization and zero-padding do not by themselves introduce semantic correspondence. In B (frozen--projected), the shared projection already pulls same-class samples closer together across modalities, although several classes remain only partially separated. In C (adapted--raw), encoder adaptation visibly increases color organization within each modality-specific region relative to A, indicating that composer-discriminative information has been introduced into the raw representation, but this information is still confined to modality-specific subspaces. In D (adapted--projected), class clusters become clearly recognizable--for instance, the orange, green, purple, brown, and red classes each form a broadly contiguous group--while still exhibiting internal subdivision by marker shape. 

\subsubsection*{\triscore{} instrument: a redundant case}

\begin{figure}[ht!]
    \centering
    \includegraphics[width=\textwidth]{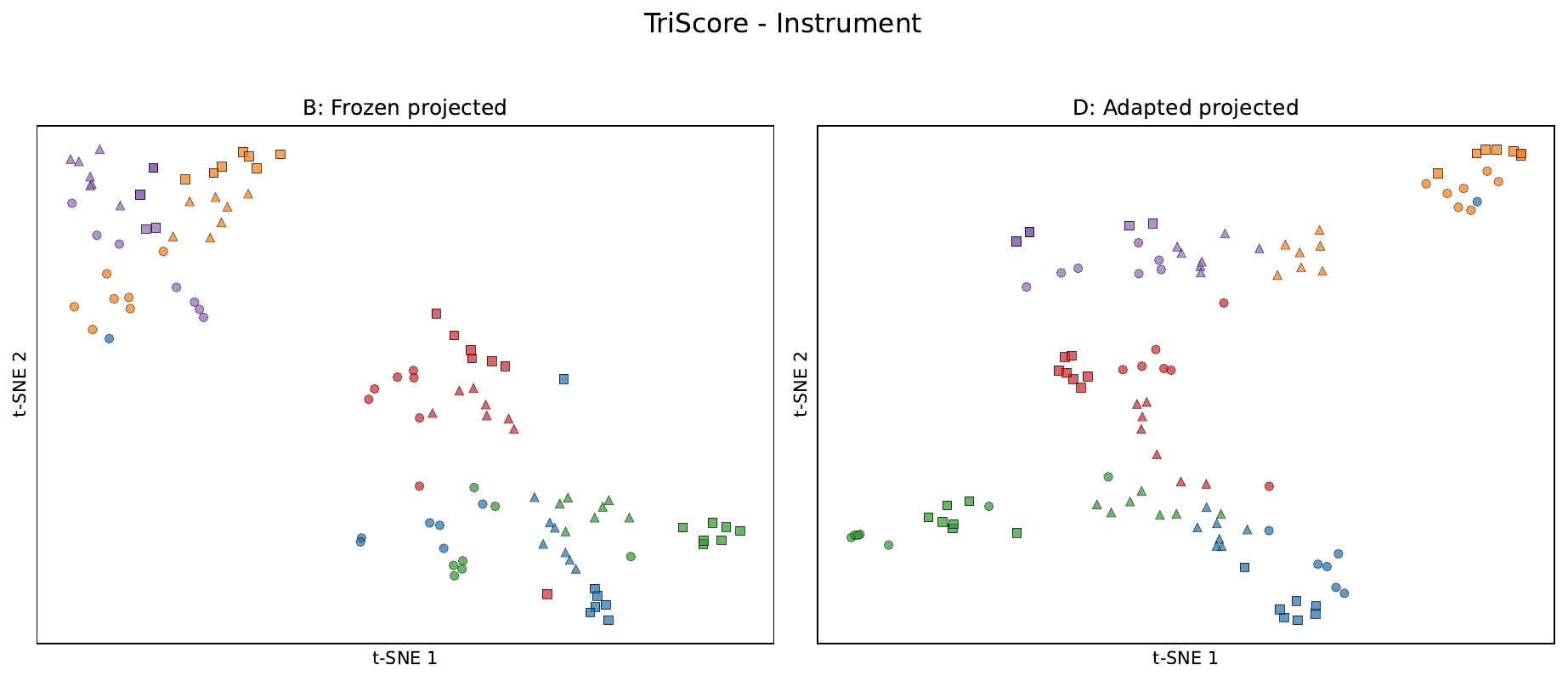}
    \caption{t-SNE visualization of the frozen--projected $((\phi_0, \mathcal{H})$and adapted--projected $((\phi_{adapt}, z)$conditions for \triscore{} instrument. Colors indicate instrument class; marker shapes indicate modality.}
    \label{fig:tsne_triscore_instrument}
\end{figure}

Figure~\ref{fig:tsne_triscore_instrument} contrasts the frozen--projected $(\phi_0, \mathcal{H})$and adapted--projected $(\phi_{adapt}, \mathcal{H})$conditions for \triscore{} instrument. Both panels display comparably well-organized class clusters, and in both, some classes remain split into modality-specific subgroups rather than forming a single unified cluster. Unlike the composer case, condition $(\phi_{adapt}, \mathcal{H})$does not show a visually stronger class organization than $(\phi_0, \mathcal{H}))$; if anything, some classes appear slightly more fragmented in $(\phi_{adapt}, \mathcal{H}))$. This qualitative similarity is consistent with the redundant interaction reported in Table~\ref{tab:ablation_full}, where the adapted--projected condition (70.03) does not improve over the frozen--projected condition (74.39), a decrease of 4.36~pp. The visualization suggests that the shared projection alone already recovers most of the instrument-discriminative structure available in the frozen representations, so that the additional structure introduced by encoder adaptation is largely redundant with what projection already provides. This is compatible with the possibility that t-SNE surfaces some of this structure in two dimensions while still altering relevant high-dimensional neighborhood relationships that determine kNN performance.

\subsubsection*{\gls{mtd} composer: an interfering case}

\begin{figure}[ht!]
    \centering
    \includegraphics[width=\textwidth]{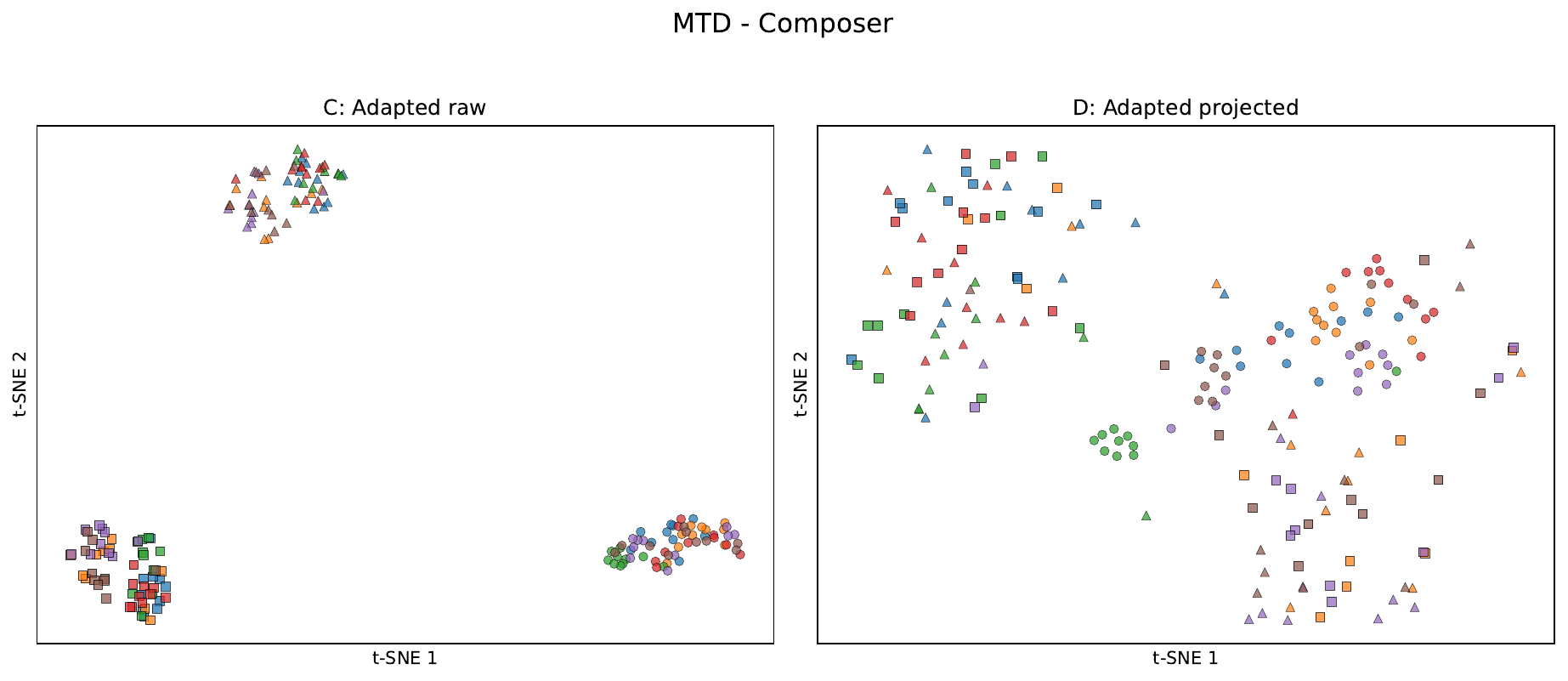}
    \caption{t-SNE visualization of the adapted--raw $(\phi_{adapt}, \mathcal{Z})$ and adapted--projected $(\phi_{adapt}, \mathcal{Z})$ conditions for \gls{mtd} composer. Colors indicate composer class; marker shapes indicate modality.}
    \label{fig:tsne_mtd_composer}
\end{figure}

Figure~\ref{fig:tsne_mtd_composer} shows the adapted--raw $(\phi_{adapt}, \mathcal{Z})$and adapted--projected $(\phi_{adapt}, \mathcal{Z})$conditions for \gls{mtd} composer, the target with the strongest interference effect in Table~\ref{tab:ablation_full}. In C, three large regions dominated by modality are clearly visible, and within each region, class colors remain substantially mixed, indicating that composer-related information is weak relative to modality identity even after adaptation. In D, some local groupings by color emerge--for example, small clusters of green, red, or purple points--but global class separation remains poor, and markers continue to segregate into modality-specific regions rather than forming unified, mixed-modality class clusters. This is consistent with the partial interference reported in Table~\ref{tab:ablation_full}, where projection reduces performance relative to the adapted--raw condition (11.41 to 8.58 macro F1, $F1_{h}^{\phi_{adapt}}-F1_{z}^{\phi_{adapt}}=-2.83$~pp): the category-supervised projection reorganizes the space but does not preserve the composer-discriminative structure already present in C. This pattern is consistent with the structural difficulty of \gls{mtd} noted in Section~\ref{sec:materials_methods:datasets:mtd}, where the visual and symbolic modalities describe short monophonic themes while the corresponding audio may carry additional harmonic, timbral, and contextual information not shared across modalities.

\subsubsection*{Overall visual pattern}

To complement the quantitative factorial evaluation (Section~\ref{sec:results:final_impact}), we additionally visualize a small number of representative cases using t-SNE~\cite{vandermaaten2008}. For each selected dataset--target combination, a balanced subset of classes and modality-matched samples is drawn, and PCA followed by t-SNE is applied \emph{independently} to each condition. Because each panel is produced through an independent fit, absolute coordinates, distances, and cluster compactness are not comparable across panels; only the relative organization of color (class) and marker shape (modality) within a given panel is meaningful. These visualizations are intended as qualitative, complementary evidence and are not used to draw quantitative conclusions, which remain based on macro F1, cross-modal exclusion, and neighborhood-composition analyses.

Across the three cases, class-related color organization and residual modality-related shape organization coexist to varying degrees, reflecting that the shared projection is trained with category-level supervision rather than an explicit objective that pulls corresponding cross-modal triplets together. This is in line with the neighborhood-composition results of Section~\ref{sec:results:analysis}, where same-modality neighbor proportions exceed 93\% in several projected spaces despite class-level coordination. Because each panel in Figures~\ref{fig:tsne_triscore_composer}--\ref{fig:tsne_mtd_composer} is produced through an independent PCA+t-SNE fit on a small, class-balanced subset, these visualizations should be read as illustrative support for the factorial and neighborhood-composition analyses rather than as an independent quantitative result.

\subsection{Performance of the final projected classification spaces}

Table~\ref{tab:projected_macro_by_modality} provides a modality-specific view of condition $(\phi_{adapt}, \mathcal{H}))$, the final adapted-and-projected representation. Macro F1 is reported separately for audio, image, and symbolic score queries using a combined gallery containing all modalities.

\begin{table}[ht]
    \caption{Classification performance in the task-specific projected spaces condition $(\phi_{adapt}, \mathcal{H}))$. Scores are macro F1 values in \gls{pp}. A separate projected space is evaluated for each dataset and target label. The gallery contains all modalities, and columns indicate the query modality.}
    \label{tab:projected_macro_by_modality}
    \centering
    \resizebox{.9\textwidth}{!}{%
    \begin{tabular}{llcccc}
    \toprule
    \textbf{Dataset} & \textbf{Label} & \textbf{Audio query} & \textbf{Image query} & \textbf{Score query} & \textbf{Avg.} \\
    \midrule
    \multirow{2}{*}{\triscore{}}
        & Composer    & 83.04 & 47.22 & 94.70 & 74.99 \\
        & Instrument  & 80.40 & 47.24 & 82.46 & 70.03 \\
    \midrule
    \multirow{3}{*}{\gls{mtd}}
        & Instrument  & 29.11 &  9.68 & 11.01 & 16.60 \\
        & Ensemble    & 27.36 & 10.80 & 10.73 & 16.29 \\
        & Composer    & 10.03 &  7.17 &  8.54 &  8.58 \\
    \bottomrule
    \end{tabular}%
    }
\end{table}

On \triscore{}, symbolic score queries achieve the strongest result for composer classification at 94.70 macro F1, followed by audio at 83.04. Image queries are substantially weaker at 47.22. Instrument classification follows a similar pattern: audio and score queries reach 80.40 and 82.46, respectively, whereas image queries reach 47.24.

The final projected spaces therefore remain asymmetric across query modalities even when their average performance is high. In particular, image is consistently the weakest query modality for both \triscore{} classification targets.

On \gls{mtd}, audio queries dominate instrument and ensemble classification, reaching 29.11 and 27.36 macro F1. Image and score queries remain near or below 11.01. For composer, the differences among query modalities are smaller, but only because all three perform poorly: audio reaches 10.03, score 8.54, and image 7.17.

This pattern is consistent with the informational content of the dataset. Audio contains direct timbral and orchestral evidence relevant to instrument and ensemble, whereas the short monophonic visual and symbolic themes omit much of this information. Composer-related discrimination remains weak across all three modalities in the final projected space.

\subsection{Piece-level retrieval with the final title representation}

The preceding factorial analysis evaluates whether unimodal adaptation transfers to category-level multimodal classification. We now address a complementary question: whether the final title-specific adapted-and-projected representation supports practical piece-level retrieval. This experiment evaluates condition $(\phi_{adapt}, \mathcal{H})$only. It therefore characterizes the retrieval capability of the complete system and does not isolate the individual contributions of encoder adaptation and shared projection.

Given a query, gallery samples are ranked according to their distance in the title-specific projected space. Retrieval is considered correct when at least one of the first $N$ samples shares the same piece-title label as the query. Table~\ref{tab:piece_retrieval_title} reports Top-5 and Top-10 accuracy.

\begin{table}[H]
    \caption{Piece-level retrieval on \triscore{} using the title label. Scores are Top-$N$ accuracies in \gls{pp} in the title-specific projected space.}
    \label{tab:piece_retrieval_title}
    \centering
    \resizebox{.75\textwidth}{!}{%
    \begin{tabular}{lcccc}
    \toprule
    \textbf{Metric} & \textbf{Audio query} & \textbf{Image query} & \textbf{Score query} & \textbf{Avg.} \\
    \midrule
    Top-5  & 61.62 & 34.59 & 69.73 & 55.32 \\
    Top-10 & 69.19 & 45.95 & 78.92 & 64.68 \\
    \bottomrule
    \end{tabular}%
    }
\end{table}

The final representation supports meaningful piece-level retrieval, with average Top-5 and Top-10 accuracies of 55.32 and 64.68, respectively. Performance nevertheless differs substantially across query modalities.

Symbolic score queries produce the strongest retrieval results, reaching 69.73 Top-5 and 78.92 Top-10 accuracy. Audio queries follow at 61.62 and 69.19. Image queries are considerably weaker, reaching 34.59 Top-5 and 45.95 Top-10.

The complete title-specific system therefore organizes excerpts from the same musical work sufficiently closely to support piece identification, particularly from symbolic and audio queries.

\subsection{Cross-modal exclusion and neighborhood analysis}
\label{sec:results:analysis}

The combined-gallery results do not by themselves demonstrate local cross-modal alignment. A query may be classified or retrieved successfully because its nearest neighbors belong predominantly to the same modality. We therefore evaluate cross-modal exclusion and analyze the modality composition of local neighborhoods in the final projected spaces.

\subsubsection{Cross-modal exclusion classification}

For each query modality, all gallery samples from that modality are removed, forcing classification to rely exclusively on the other two modalities. Table~\ref{tab:cross_modal_exclusion_macro} reports test macro F1 under this setting for the final adapted-and-projected classification spaces.

Cross-modal exclusion is reported for composer and instrument on \triscore{} and for all three category-level targets on \gls{mtd}. Title is excluded because the present retrieval evaluation does not include an equivalent cross-modal exclusion protocol.

\begin{table}[H]
    \caption{Cross-modal exclusion classification in the task-specific projected spaces. Scores are macro F1 values in \gls{pp}. For each query modality, samples from that same modality are removed from the gallery.}
    \label{tab:cross_modal_exclusion_macro}
    \centering
    \resizebox{.9\textwidth}{!}{%
    \begin{tabular}{llcccc}
    \toprule
    \textbf{Dataset} & \textbf{Label} & \textbf{Audio query} & \textbf{Image query} & \textbf{Score query} & \textbf{Avg.} \\
    \midrule
    \multirow{2}{*}{\triscore{}}
        & Composer    & 77.47 & 17.54 & 89.25 & 61.42 \\
        & Instrument  & 67.77 & 35.97 & 68.49 & 57.41 \\
    \midrule
    \multirow{3}{*}{\gls{mtd}}
        & Instrument  & 14.39 &  8.90 & 22.12 & 15.14 \\
        & Ensemble    & 20.30 & 10.30 & 14.75 & 15.11 \\
        & Composer    &  2.49 &  7.69 &  8.00 &  6.06 \\
    \bottomrule
    \end{tabular}%
    }
\end{table}

On \triscore{}, composer classification remains strong for audio and score queries after removing same-modality gallery samples, reaching 77.47 and 89.25 macro F1, respectively. This indicates that composer-related information is represented in a sufficiently compatible form for these queries to use neighbors from the other modalities.

Image queries behave differently. Their composer macro F1 decreases from 47.22 with the combined gallery to 17.54 under cross-modal exclusion. Instrument shows the same qualitative pattern: audio and score queries retain 67.77 and 68.49 macro F1, while image queries reach 35.97. Image is therefore the weakest cross-modal query modality on \triscore{} for both classification targets.

On \gls{mtd}, cross-modal exclusion performance remains low for all labels. Instrument and ensemble reach average macro F1 values of 15.14 and 15.11, while composer falls to 6.06. The particularly weak composer result is consistent with the factorial analysis, where the projection removed 2.83~pp from the adapted raw representation. Together, these results indicate that the final composer space does not establish a robust cross-modal category structure on \gls{mtd}.

\begin{figure}[ht!]
    \centering
    \begin{subfigure}[t]{.9\textwidth}
        \centering
        \includegraphics[width=\textwidth]{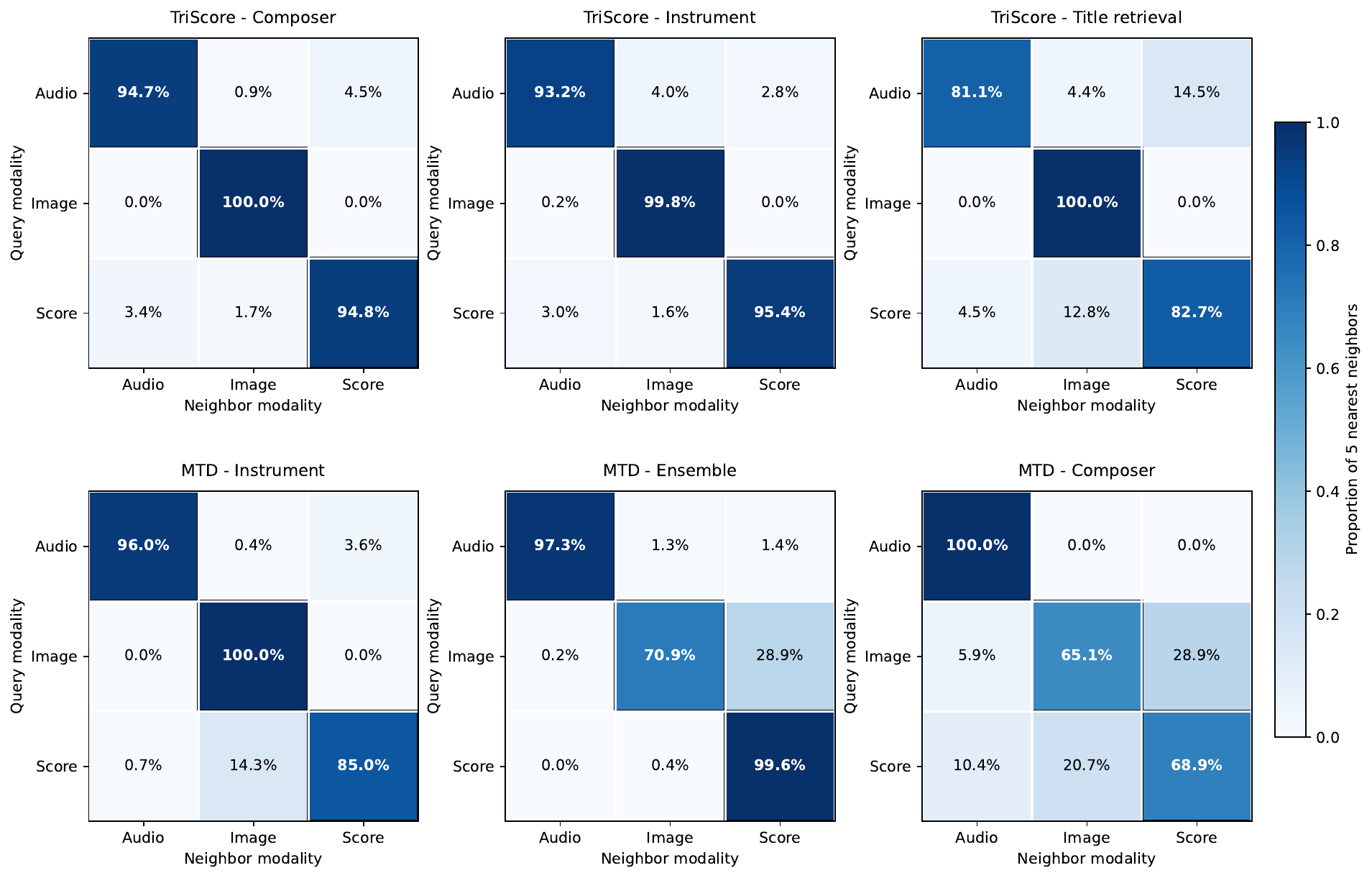}
        \caption{Full gallery (same-modality on the diagonal).}
        \label{fig:neighbor_composition_full}
    \end{subfigure}
    \vspace{0.6cm}
    \begin{subfigure}[t]{.9\textwidth}
        \centering
        \includegraphics[width=\textwidth]{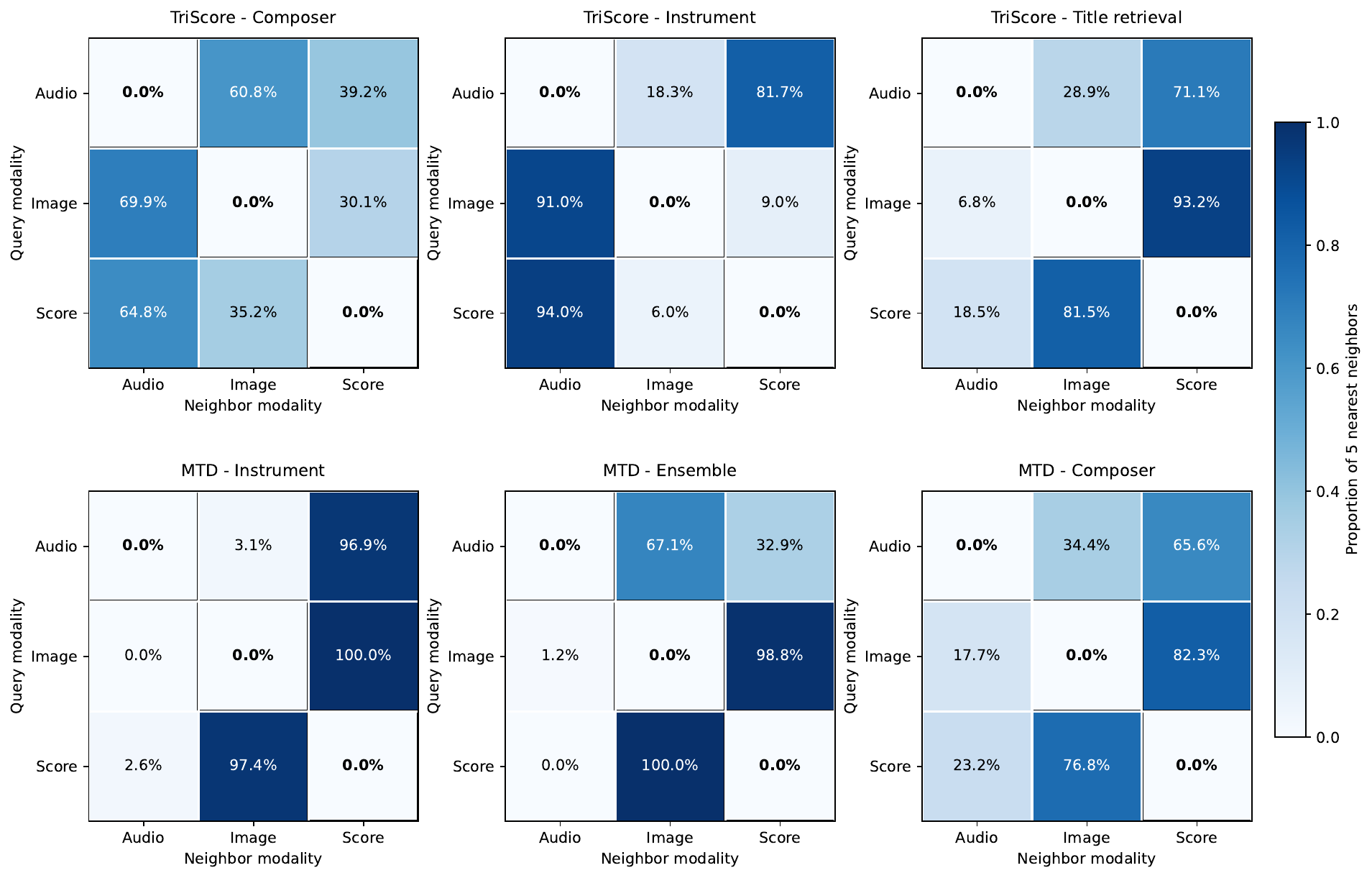}
        \caption{Cross-modal exclusion (query modality removed from the gallery; diagonal excluded by construction).}
        \label{fig:neighbor_composition_exclusion}
    \end{subfigure}
    \caption{Neighborhood modality composition in the task-specific projected spaces. Each heatmap reports, for a given query modality (rows), the proportion of its five nearest neighbors belonging to each neighbor modality (columns). (a) All three modalities are available in the gallery; the diagonal is the same-modality proportion. (b) The query modality is excluded from the gallery; the diagonal cells are not defined and are shown in grey.}
    \label{fig:neighbor_composition}
\end{figure}

\subsubsection{Neighborhood modality composition}

To further examine local geometry, we compute the modality composition of the five nearest neighbors of every query. Figure~\ref{fig:neighbor_composition} reports two complementary quantities as 3$\times$3 heatmaps, one panel per dataset--target combination, with rows indicating the query modality and columns indicating the neighbor modality. Figure~\ref{fig:neighbor_composition_full} shows the \emph{same-modality} composition: the percentage of the five nearest neighbors belonging to each modality when all three modalities are available in the gallery, with the diagonal reporting the proportion of neighbors that share the query modality. Figure~\ref{fig:neighbor_composition_exclusion} shows the composition after the query modality is removed from the gallery; the diagonal is undefined by construction in this condition and is shown as an excluded cell, while the two off-diagonal columns report how the remaining neighbors are distributed between the other two modalities.

Because this analysis concerns representation geometry rather than classification performance, it includes the composer and instrument classification spaces and the title-specific retrieval space for \triscore{}.

The full-gallery analysis shows that the final task-specific projected spaces remain strongly organized by modality. On \triscore{}, same-modality proportions exceed 93\% for every composer and instrument query type. Image queries reach 100\% same-modality neighbors for composer and title, and 99.83\% for instrument. The title representation is somewhat less segregated for audio and score queries, at 81.08\% and 82.70\%, but its local neighborhoods are still predominantly unimodal.

The effect is similarly strong on \gls{mtd}. Same-modality proportions reach 96\% or more for several instrument and ensemble conditions, including 100\% for image queries in instrument and audio queries in composer. These results show that a projection shared across modalities does not necessarily produce locally mixed neighborhoods.

When the query modality is excluded (Figure~\ref{fig:neighbor_composition_exclusion}), the remaining neighbors exhibit structured but target-dependent pairings. On \triscore{}, the title space links audio and image queries primarily to symbolic scores, with proportions of 71.14\% and 93.19\%, respectively. Score queries instead retrieve image neighbors most frequently, at 81.51\%.

The composer space produces a different pattern. Audio queries retrieve images most frequently, while image and score queries retrieve audio. For instrument, audio and score form the strongest pairing, whereas image queries also retrieve audio predominantly. Thus, even within the same dataset, different supervision targets induce different cross-modal neighborhood structures.

On \gls{mtd}, image and symbolic score frequently form a dominant pair. For instrument, image queries retrieve score neighbors exclusively, while score queries retrieve image neighbors in 97.40\% of cases. A similar relationship appears for ensemble. Audio does not exhibit a single stable partner across targets, reflecting the informational difference between the richer audio excerpts and the shorter visual and symbolic themes.

Overall, the final projected representations exhibit partial cross-modal category structure without full local modality integration. They can support classification across modalities and, in the title condition, piece-level retrieval, while their nearest-neighbor neighborhoods remain predominantly organized by input modality. This distinction reinforces the main factorial finding: improving modality-specific discrimination and learning a category-supervised shared projection do not necessarily produce instance-level cross-modal alignment.

\section{Conclusions}
\label{sec:conclusions}

This work addressed two related questions: whether parameter-efficient fine-tuning provides a competitive alternative to full fine-tuning for score-related modality encoders, and whether the resulting unimodal improvements translate into better multimodal representations. We evaluated full fine-tuning, \gls{lora}, and \gls{qlora} across nine pretrained audio, sheet music image, and symbolic score encoders on \triscore{} and \gls{mtd}. The selected configurations were subsequently assessed through a controlled $2\times2$ design crossing encoder adaptation with task-specific shared projection.

The unimodal results establish \gls{peft} as an effective strategy for adapting heterogeneous music encoders. \gls{lora} or \gls{qlora} is selected in 14 of the 18 dataset--modality--target combinations, frequently matching or outperforming full fine-tuning while optimizing substantially fewer parameters. Selected \gls{peft} configurations require approximately 0.7--1.6M trainable parameters for most audio and symbolic encoders, compared with more than 100M under full fine-tuning, while the selected PaliGemma configurations optimize 21.9M parameters. These reductions are particularly relevant in multimodal systems, where several modality- and target-specific encoder variants must be trained and maintained. Nevertheless, \gls{peft} is not uniformly superior: its effectiveness depends on the backbone, modality, dataset, and supervision target, and full fine-tuning remains preferable in several conditions.

The results also show that the pretraining domain alone does not determine downstream suitability. Music-specific \mert{} is selected for some audio targets, but Wav2Vec2 and Whisper provide the strongest representations in others. Similarly, GraphCodeBERT frequently outperforms general language encoders for notation-level symbolic scores despite being pretrained on source code. Encoder performance therefore appears to depend on the interaction among architecture, pretraining objective, input structure, and supervision target, rather than solely on whether the pretraining data are music-specific.

The factorial evaluation demonstrates that stronger unimodal representations do not uniformly produce stronger multimodal representations. Adaptation and projection are complementary for \triscore{} composer classification, where the adapted-and-projected condition reaches 74.99 macro F1 and improves by 32.90~\gls{pp} over the frozen--raw reference. For \triscore{} instrument and the \gls{mtd} instrument and ensemble tasks, the two transformations are largely redundant: each can improve the frozen representation independently, but their combination provides little additional benefit. For \gls{mtd} composer, they partially interfere, as adaptation improves the raw representation from 4.38 to 11.41 macro F1 before projection, but performance decreases to 8.58 after projection.

These findings indicate that adaptation can emphasize information that is predictive within a modality but not represented consistently across modalities. Transfer is more favorable when the modalities contain closely corresponding musical information, as in \triscore{}, and more difficult when their informational content differs, as in \gls{mtd}, where audio contains harmonic, timbral, and contextual cues absent from the monophonic visual and symbolic themes. Because the multimodal stage uses the adaptation strategy selected independently for each modality, it evaluates the transfer of the selected adaptation pipeline rather than isolating the specific multimodal effects of full fine-tuning, \gls{lora}, and \gls{qlora}.

The final adapted-and-projected title representation on \triscore{} supports piece-level retrieval, reaching average Top-5 and Top-10 accuracies of 55.32 and 64.68, respectively. Cross-modal exclusion further shows that the projected spaces can support classification using other modalities, particularly for audio and symbolic score queries on \triscore{}. However, nearest-neighbor neighborhoods remain predominantly organized by input modality. Category-level semantic coordination can therefore emerge without complete instance-level cross-modal alignment.

Overall, \gls{peft} provides a practical and competitive approach to adapting pretrained encoders for score-based multimodal music processing, but improved unimodal discrimination alone is not sufficient to guarantee better multimodal organization. Encoder adaptation and shared projection should be evaluated as separate, potentially interacting components. Future work should combine parameter-efficient adaptation with objectives that explicitly promote cross-modal proximity, including contrastive, metric-learning, and multi-positive losses based on aligned triplets or shared musical works. Modality-specific or asymmetric mappings may also help reconcile modalities whose musical content is only partially equivalent.




\bibliographystyle{model5-names} 


\bibliography{paper}

\end{document}